\documentclass[letterpaper, 10 pt, conference]{ieeeconf}

\IEEEoverridecommandlockouts                              % This command is only needed if 
\usepackage[utf8]{inputenc}
\usepackage{comment}
\usepackage{booktabs}

\usepackage{graphicx}
\usepackage{caption}
\usepackage{subcaption}
\usepackage{amsmath}
\usepackage{verbatim}
\usepackage{amsfonts}
\usepackage{multirow}
\usepackage{changepage}
\usepackage{tabularx}
\usepackage{adjustbox}
\usepackage{mathtools}
\usepackage{amssymb}
\usepackage{soul}
\usepackage{color}
\usepackage[noend]{algpseudocode}

\usepackage{color}
\usepackage{tikz}
\usepackage{flushend}
\usetikzlibrary{shapes.geometric, arrows}
\usepackage[linesnumbered,ruled]{algorithm2e}
\usepackage{multirow}
\usepackage{soul,color}
\usepackage{graphicx}
\usepackage[colorlinks]{hyperref}
\usepackage{fancyhdr}
\usepackage{comment}
\usepackage{epsfig}
\usepackage{bm}
\usepackage{float}
\usepackage{cuted}
\allowdisplaybreaks

\makeatletter
\def\algbackskip{\hskip-\ALG@thistlm}
\makeatother

\newcommand{\argmin}{\operatorname{arg min}}

\newcommand\oprocendsymbol{\hbox{$\square$}}
\newcommand\oprocend{\relax\ifmmode\else\unskip\hfill\fi\oprocendsymbol}

\newtheorem{theorem}{Theorem}

\newtheorem{lemma}{Lemma}

\newtheorem{remark}{Remark}

\renewcommand{\theenumi}{(\roman{enumi}}

\makeatletter
\let\NAT@parse\undefined
\makeatother

\renewcommand{\baselinestretch}{0.94}

\title{\LARGE \bf
Interaction-Aware Trajectory Planning for Autonomous Vehicles with Analytic Integration of Neural Networks into Model Predictive Control
}

\author{Piyush Gupta\textsuperscript{1,2*}  \hspace{1.2cm}   David Isele\textsuperscript{2}  \hspace{1.2cm}   Donggun Lee\textsuperscript{3}     \hspace{1.2cm} Sangjae Bae\textsuperscript{2} 
\thanks{
\textsuperscript{1} Michigan State University, East Lansing, MI, 48824, USA. \texttt{\{guptapi1\}@msu.edu} \ \textsuperscript{*}Corresponding Author}
\thanks{
\textsuperscript{2} Honda Research Institute, San Jose, CA, 95134, USA. \texttt{\{piyush{\_}gupta, disele, sbae\}@honda-ri.com }}
\thanks{
\textsuperscript{3} Massachusetts Institute of Technology, Cambridge, MA, 02139, USA.
\texttt{\{donggun\}@mit.edu}}
\thanks{This work has been supported in part by Honda Research Institute, USA, and NSF Award ECCS-2024649.}
}

\begin{document}

\maketitle
\thispagestyle{plain}
\pagestyle{plain}
\pagenumbering{gobble}
%%%%%%%%%%%%%%%%%%%%%%%%%%%%%%%%%%%%%%%%%%%%%%%%%%%%%%%%%%%%%%%%%%%%%%%%%%%%%%%%
\begin{abstract}

% This paper presents an interaction-aware trajectory planner for autonomous driving that can interact with the surrounding vehicles to perform complex maneuvers in a locally-optimal manner.
% Autonomous vehicles (AVs) must share the driving space with other human drivers. AVs often employ conservative motion planning strategies to ensure safety in the presence of uncertain human drivers. 
Autonomous vehicles (AVs) must share the driving space with other drivers and often employ conservative motion planning strategies to ensure safety. These conservative strategies can negatively impact AV's performance and significantly slow traffic throughput. Therefore, to avoid conservatism, we design an interaction-aware motion planner for the ego vehicle (AV) that interacts with surrounding vehicles to perform complex maneuvers in a locally optimal manner. Our planner uses a neural network-based interactive trajectory predictor and analytically integrates it with model predictive control (MPC). We solve the MPC optimization using the alternating direction method of multipliers (ADMM) and prove the algorithm's convergence. We provide an empirical study and compare our method with a baseline heuristic method.
\end{abstract}

%%%%%%%%%%%%%%%%%%%%%%%%%%%%%%%%%%%%%%%%%%%%%%%%%%%%%%%%%%%%%%%%%%%%%%%%%%%%%%%%

\section{Introduction}
Motion planning for autonomous vehicles (AVs) is a daunting task, where AVs must share the driving space with other drivers. Driving in shared spaces is inherently an interactive task, i.e., AV's actions  affect other nearby vehicles and vice versa~\cite{ulbrich2015structuring}. This interaction is evident in dense traffic scenarios where all goal-directed behavior relies on the cooperation of other drivers to achieve the desired goal. 
To predict the nearby vehicles' trajectories, AVs often rely on simple predictive models such as assuming constant speed for other vehicles~\cite{tariq2022vehicle}, treating them as bounded disturbances~\cite{gray2013robust}, or 
approximating their trajectories using a set of known trajectories~\cite{vasudevan2012safe}.
% While human drivers can predict the reactions of nearby vehicles due to their actions, AVs often rely on simple predictive models to predict the trajectory of  nearby vehicles. These models include: assuming constant speed for other vehicles~\cite{ Faizan_overtake}, treating them as bounded disturbances~\cite{gray2013robust}, or approximating their trajectories using a set of known trajectories~\cite{vasudevan2012safe}.
These models do not capture the inter-vehicle interactions in their predictions. As a result, AVs equipped with such models struggle under challenging scenarios that require interaction with  other vehicles~\cite{brito2021learning, bae2020cooperation}.

\begin{figure}
	\begin{subfigure}[b]{0.18\textwidth}
	    \centering
        \includegraphics[width=1.3\linewidth, height=1.3\linewidth, keepaspectratio]{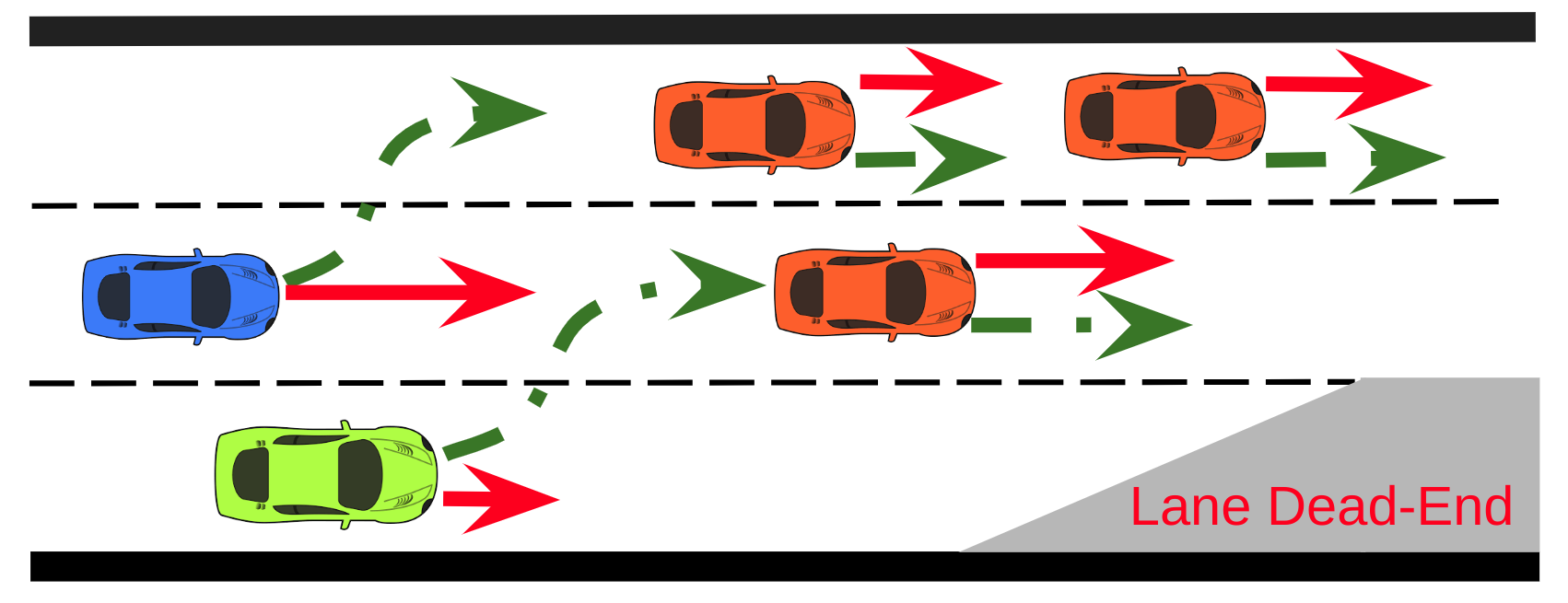}
        \caption{}
        \label{fig:scenario1}
    \end{subfigure}
    ~~~~~~~~~
	\begin{subfigure}[b]{0.18\textwidth}
	\centering
	\includegraphics[width=1.3\linewidth, height=1.3\linewidth, keepaspectratio]{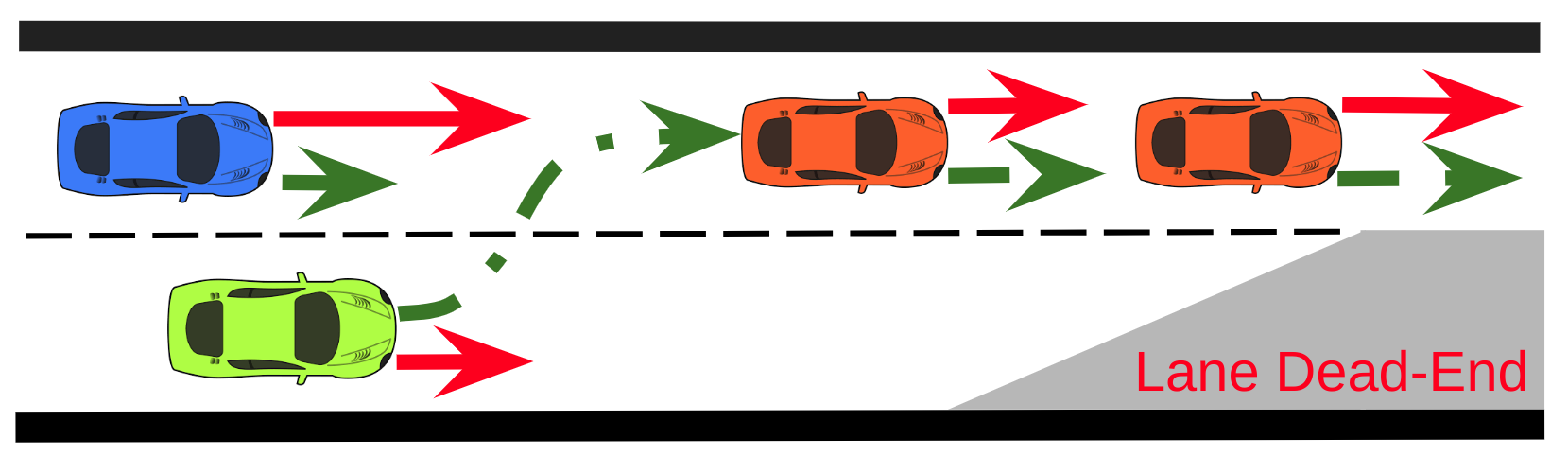}
        \caption{}
        \label{fig:scenario2}
    \end{subfigure}

    \caption{\footnotesize{Dense traffic scenarios where the ego vehicle (green) intends to merge to the left lane. The red and green trajectories show the nominal (conservative) and interaction-aware trajectories for the ego vehicles, respectively, and correspondingly, their impact on the other vehicles. Due to interaction with the ego vehicle (green trajectory), in scenario (a), the blue vehicle switches lanes, and in scenario (b), the blue vehicle slows down to create space for the ego vehicle to perform a safe lane-change maneuver.    % Due to interaction with the ego vehicle (green trajectory), in scenario (a), the blue vehicle switches lanes to make space for the ego vehicle. Similarly, in scenario (b), the blue vehicle slows down 
    % % and the orange vehicles speed up
    % to create space for the ego vehicle to perform a safe lane-change maneuver. 
    }} 
    % \di{ It's unusual that the orange cars would change their behavior for the cars behind them}
    % \pg{ Fixed: Made red arrows for orange vehicles longer and removed the text about orange vehicles speeding up.}
%\vspace{-0.1in}
    \label{fig:scenarios}
\end{figure}

% Today's AVs are equipped with high-quality sensors and advanced algorithms for perception, localization, lane keeping, etc. However,
AVs can be overly defensive and opaque when interacting with other drivers~\cite{sadigh2016planning}, as they often rely on decoupled prediction and planning techniques \cite{burger2020interaction}. The prediction module anticipates the trajectories of other vehicles, and the planning module uses this information to find a collision-free path. As a result of this decoupling, AVs tend to be conservative and treat other vehicles as dynamic obstacles, resulting in a lack of cooperation \cite{sheng2022cooperation}. 
% When interacting with other drivers, AVs tend to be overly defensive and obliviously opaque~\cite{sadigh2016planning}. This is because they use decoupled prediction and planning techniques~\cite{burger2020interaction}. The prediction module predicts the trajectories of other vehicles, and the planning module uses these predictions to find a collision-free trajectory. This decoupling results in a conservative AV~\cite{sheng2022cooperation} that treats other vehicles as dynamic obstacles and avoids their trajectories.
Figure~\ref{fig:scenarios} shows two scenarios in which the ego vehicle intends to merge into the left lane, but the inter-vehicle gaps are too narrow.
% for the lane change maneuver. 
In such scenarios, conservative AVs with decoupled prediction and planning are forced to wait for a long duration. In contrast, we propose an interaction-aware AV that can open up a gap for itself by negotiating with other agents, i.e., by nudging them to either switch lanes (Fig.~\ref{fig:scenario1}) or change speeds (Fig.~\ref{fig:scenario2}). 
% \di{The first two paragraphs are a nice introduction, but a bit verbose. Try reduce so that the introduction mostly fits on the first page.} 

Reinforcement learning (RL) techniques~\cite{9992857} have been used to learn control policies under interactive or unknown environments. For example, adversarial RL is designed to reach the desired goal under uncertain conditions~\cite{9991836}, and a model-free RL agent is developed for lane-changing control in dense traffic environments~\cite{saxena2020driving}. However, these RL methods are not yet appropriate for safety-critical AVs due to their low interpretability and reliability.
% applications such as autonomous driving.
% \blue{The ego vehicle must negotiate with the other agents to open a gap for itself to merge into.
% These challenging conditions must be addressed if the AVs have to co-exist with the other human drivers and share the same road. Therefore, in order for AVs to co-exist with the other human drivers and operate in challenging environments, it is critical for them to be interaction-aware and utilize the interactions for its trajectory planning. 

Designing interaction-aware planners presents a significant challenge, as predicting the reactions of surrounding vehicles to the ego vehicle's actions is complex and non-trivial. Data-driven approaches, such as those using recurrent neural network architectures, have been effective in capturing the complex interactive behaviors of agents \cite{choi2019drogon, gupta2018social}, especially in predicting driver behavior with high accuracy and computational efficiency \cite{gupta2018social}. Therefore, it is desirable to utilize these data-driven methods to predict other vehicles' interactive behavior while maintaining safety through rigorous control theory and established vehicle dynamics models.

We propose a model predictive control (MPC) based motion planner that incorporates AV’s decision and surrounding vehicles’ interactive behaviors into safety constraints to perform complex maneuvers. In particular, we provide a mathematical formulation for integrating the neural network's predictions in the MPC controller and provide methods to obtain an (locally) optimal solution. However, the neural network integration and non-linear system dynamics make the optimization highly non-convex and challenging to solve analytically. Thus, prior efforts~\cite{bae2020cooperation, bae2022lane, sheng2022cooperation} that integrate neural network prediction into MPC are numerical in nature and rely on heuristic algorithms to generate a finite set of trajectory candidates. In~\cite{bae2020cooperation} and~\cite{bae2022lane}, the authors generate these candidates by random sampling of control trajectories, and by generating spiral curves from source to target lane, respectively. In~\cite{sheng2022cooperation}, the authors utilize a predefined set of reference trajectory candidates. Instead of solving the optimization, these approaches evaluate the cost of each candidate and choose the minimum cost trajectory that satisfies the safety constraint. Optimality is therefore restricted to trajectory candidates only, and the planner's performance depends on the heuristic algorithm design. 
% \di{consider removing for space} \st{Moreover, these methods often fail to generate viable trajectories in complex scenarios due to limited obstacle-free space and limited candidates generated by the heuristic algorithm.} 
In contrast to these prior efforts, we avoid heuristics, detail a proper formalization, and solve the optimization with provable optimality. The optimal solution provides key insights to design better planners and can be leveraged to compare trajectories obtained by other heuristic methods.

The major contributions of this work are twofold: (i)
we reformulate a highly complex MPC problem with a non-convex neural network and non-linear system dynamics, and systematically solve it using the Alternating Direction Method of Multiplier (ADMM)~\cite{boyd2011distributed} with generic assumptions (Section~\ref{Sec:ADMM-formulation}), and (ii) we investigate the mathematical properties of the ADMM algorithm for solving the MPC with an integrated neural network. Specifically,  we provide sufficient conditions on the neural network such that the ADMM algorithm in non-convex optimization converges to a local optimum (Section~\ref{Convergence of ADMM}). It is one of the first attempts in the literature toward provable mathematical guarantees for a neural network-integrated MPC.

\section{Problem Formulation and Controller Design}\label{Sec:Mathematical-formulation}
% \bae{The major part of section A-C can be simplified by referring to the NNMPC papers. Again, our contribution is not the MPC formulation with NN (it's done by the NNMPC papers), but ``reformulation'' (in compact form) for ADMM.}
We design an MPC controller that leverages interactive behaviors of surrounding $N \in \mathbb{N}$ vehicles conditioned on the ego vehicle's future actions. The key to leverage interactions is to integrate a neural network and interactively update controls with step-size $\Delta t \in \mathbb{R}_{>0}$ based on its inference (i.e., predicted positions during updates). This section further details the mathematical formulation of the MPC with the neural network. 

\begin{comment}
We formulate an interaction-aware discrete-time MPC controller with step-size $\Delta t \in \mathbb{R}_{>0}$ for an ego vehicle that is surrounded by $N \in \mathbb{N}$ nearby vehicles.
The ego vehicle predicts the interactive trajectory of all the nearby vehicles using a neural network and finds an optimal trajectory that is safe with respect to (w.r.t) neural network's predictions. 
We integrate the neural network predictions in the MPC optimization and  solve it to obtain the optimal solution. The MPC controller plans a low-level trajectory for the ego vehicle based on the high-level reference commands provided by a high-level motion planner~\cite{paden2016survey}. 
\end{comment} 
% \red{Improve this paragraph, schematic of the MPC} \red{Define w.r.t}

Motivated by \cite{kong2015kinematic}, we use bicycle kinematics. The corresponding states are $[$xy-coordinates, heading angle, speed$]$ denoted by $z(\tau) = [x(\tau), y(\tau), \psi(\tau), v(\tau)]^{\top}$ for all $\tau \in \{0,\ldots,T_p\}$ and the control inputs are $[$acceleration, steering angle$]$ denoted by $[a(\tau), \delta(\tau)]$ for all $\tau \in \{0,\ldots,T_p-1\}$ with the planning horizon $T_p \in \mathbb{N}$. For brevity, let $g(\tau)$ denote any general function $g(\cdot)$ at discrete time-step $\tau \in \mathbb{Z}_{\ge 0}$ with respect to (w.r.t) time $t$, i.e. $g(\tau) \equiv g(t+\tau \Delta t)$. 

Then, at any time $t$, we solve the MPC to obtain the optimal control trajectories  $\boldsymbol{\Delta^\ast} (t) \in \mathcal{D}\subset \mathbb{R}^{T_p}$ and $\boldsymbol{\alpha^\ast} (t) \in \mathcal{A}\subset \mathbb{R}^{T_p}$, and corresponding optimal state trajectory $\boldsymbol{Z}^\ast(t) \in \mathcal{Z} \subset \mathbb{R}^{4T_p}$, where:
\begin{align}
    \boldsymbol{\Delta^\ast} (t) &= \begin{bmatrix}\delta^\ast(0),  \ldots,  \delta^\ast(T_p-1) \end{bmatrix}^\top,  &\mathcal{D}= [\delta_{min}, \delta_{max}], \nonumber\\ 
    \boldsymbol{\alpha^\ast} (t) &= \begin{bmatrix} a^\ast(0),  \ldots,  a^\ast(T_p-1) \end{bmatrix}^\top,   &\mathcal{A}= [a_{min}, a_{max}], \nonumber\\ 
    \boldsymbol{Z}^\ast(t) &= \begin{bmatrix}z^\ast(1), \ldots ,  z^\ast(T_p)\end{bmatrix}^\top, &\mathcal{Z}=[z_{min}, z_{max}].\nonumber
    % \mathcal{D} &= [\delta_{min}, \delta_{max}] \subset \mathbb{R}^{T_p}, \nonumber\\
    % \mathcal{A} &= [a_{min}, a_{max}] \subset \mathbb{R}^{T_p}, \nonumber\\
    % \mathcal{Z} &= [z_{min}, z_{max}] \subset \mathbb{R}^{4T_p}.\nonumber
\end{align}

\subsection{Objective function}

The controller's objective is to move from the current lane to the desired lane as soon as possible while minimizing control effort and ensuring safety and smoothness.
% provided by the high-level planner while guaranteeing collision avoidance. 
% We assume that it is preferred to transition to the desired lane as soon as possible while ensuring safety and smoothness. 
Let $x^{\text{ref}}$ denote the maximum longitude coordinate until when the ego must transition to the target lane.  Let $\|\cdot\|$ denote the Euclidean norm. 
% At any time $t$, let $\boldsymbol{\Delta} = \begin{bmatrix}\delta(0)&\cdots&\delta(T_p-1)\end{bmatrix}^\top, \boldsymbol{\alpha} = \begin{bmatrix}a(0)&\cdots&a(T_p-1)\end{bmatrix}^\top$, and $\boldsymbol{Z} = \begin{bmatrix}z(1) & \ldots & z(T_p)\end{bmatrix}^\top,$ be the steering trajectory, acceleration trajectory, and state trajectory, respectively, over the planning horizon length $T_p$.
For $x < x^{\text{ref}}$, we utilize the following objective (cost) function $J(\boldsymbol{\Delta}(t),\boldsymbol{\alpha}(t), \boldsymbol{Z}(t))$ similar to~\cite{bae2022lane}:
\begin{align*}
    J&=\sum_{\tau=1}^{T_p}\lambda_{div} \|y(\tau)-y^{\text{ref}}\|^2+ \sum_{\tau=1}^{T_p}\lambda_{v}\|v(\tau)-v^{\text{ref}}\|^2 \ \text{(error)}\\
     &+\sum_{\tau=0}^{T_p-1}\lambda_{\delta}\|\delta(\tau)\|^2 +\sum_{\tau=0}^{T_p-1}\lambda_{a}\|a(\tau)\|^2 \ \ \ \ \ \ \text{(control effort)}\\
     &+\sum_{\tau=0}^{T_p-1}\lambda_{\Delta\delta}\|\delta(\tau)-\delta(\tau-1)\|^2   \ \ \ \ \ \ \ \ \ \ \ \ \  \text{(steering rate)}\\
     &+\sum_{\tau=0}^{T_p-1}\lambda_{\Delta a}\|a(\tau)-a(\tau-1)\|^2, \ \ \ \ \ \ \ \ \  \ \ \ \  \ \ \ \ \ \ \ \ \text{(jerk)}
\end{align*}
where $\boldsymbol{\Delta}(t) \in \mathcal{D}$, $\boldsymbol{\alpha}(t) \in \mathcal{A}$, and $\boldsymbol{Z}(t) \in \mathcal{Z}$ are the planned steering, acceleration, and state trajectories, respectively.
% $z(\tau) = \begin{bmatrix}x(\tau)&y(\tau)&\psi(\tau)&v(\tau)\end{bmatrix}^\top$ and $u(\tau ) = \begin{bmatrix}\delta(\tau) & a(\tau)\end{bmatrix}^\top$ are the predicted states and control inputs at time $\tau$ (w.r.t time $t$) based on the measurements and predictions at time $t$, respectively.
% $x^\text{ref}$ is the longitude coordinate (along the road) of the road-end and 
$y^\text{ref} \in \mathbb{R}$ and $v^\text{ref} \in \mathbb{R}_{>0}$ are the reference latitude coordinate of the desired lane and desired velocity, respectively, provided by a high-level planner~\cite{paden2016survey}.
For a detailed description of each term, we refer the interested readers to~\cite{bae2022lane}.
% Each penalty term is regularized with a positive coefficient $\lambda_{div}$, $\lambda_{\delta}$, $\lambda_v$, $\lambda_{a}$, $\lambda_{\Delta \delta}$, and $\lambda_{\Delta a} \in \mathbb{R}_{>0}$, respectively.
% The terms in  \eqref{eq:obj_states} penalizes the lateral cross-track error w.r.t $y^\text{ref}$ and the error in velocity w.r.t $v^\text{ref}$, respectively. The terms in \eqref{eq:obj_control} penalize the control effort of the steering angle and acceleration, respectively. The terms \eqref{eq:obj_front_wheel_angle_jerk} and \eqref{eq:obj_jerk} penalize the steering rate and jerk, respectively, for the drive comfort. 

\subsection{State Dynamics}
% We utilize the discrete-time nonlinear kinematic bicycle model~\cite{kong2015kinematic} $z(\tau +1) = f(\delta(\tau), a(\tau), z(\tau))$ to represent the vehicle dynamics which \di{we linearize as...} \di{since we use the linearized version, we could remove the non-linear version for space} is re-written as:
% \begin{align}\label{eq:non-linear dynamics}
% &\begin{cases}
%     x(\tau+1)= x(\tau) + \Delta t \times v(\tau)\cos(\psi(\tau)+\beta(\tau)),\\
%     y(\tau+1)=y(\tau)+ \Delta t \times v(\tau)\sin(\psi(\tau)+\beta(\tau)),\\
%     \psi(\tau+1)= \psi(\tau) + \Delta t \times \frac{v(\tau)}{l_r}\sin(\beta(\tau)),\\
%     v(\tau+1)= v(\tau) + \Delta t \times a(\tau), \ \ \text{where}\end{cases} \\
%     &\beta(\tau)=\tan^{-1}\left(\frac{l_r}{l_f+l_r}\tan(\delta(\tau))\right), \nonumber
% \end{align}
% $l_f$ and $l_r$ indicate the distance from the center of the car to the front axle and to the rear axle, respectively.

Let $\tilde{\delta}, \tilde{a}$ and $\tilde{z}$ be the last observed steering input, acceleration input and state of the ego vehicle, respectively. At any time $t$, we linearly approximate the discrete-time kinematic bicycle model~\cite{kong2015kinematic} of the form $z(\tau +1) = f(\delta(\tau), a(\tau), z(\tau))$ 
% system dynamics $f(\delta(\tau), a(\tau), z(\tau))$ around $(\tilde{\delta}, \tilde{a}, \tilde{z})$
about $(\tilde{\delta}, \tilde{a}, \tilde{z})$ to obtain the equality constraints for the optimization problem. We have
% \di{can we clean up the formatting on this?}
\begin{align}
f(\delta(\tau), a(\tau), z(\tau)) &\approx
% f(\tilde{\delta}, \tilde{a}, \tilde{z}) +\frac{ \partial f }{\partial \delta}\Big|_{(\tilde{\delta}, \tilde{a}, \tilde{z})}(\delta(\tau)-\tilde{\delta})  + \nonumber \\
% \frac{ \partial f}{\partial a}\Big|_{(\tilde{\delta}, \tilde{a}, \tilde{z})}&(a(\tau)-\tilde{a}) + \frac{ \partial f }{\partial z}\Big|_{(\tilde{\delta}, \tilde{a}, \tilde{z})}(z(\tau)-\tilde{z}) \\
% &:=
\tilde{A}\delta(\tau) + \tilde{B} a(\tau) + \tilde{C} z(\tau) + \tilde{D},
\end{align}
where $\tilde{A} \in \mathbb{R}^{4}, \tilde{B} \in \mathbb{R}^{4}, \tilde{C} \in  \mathbb{R}^{4 \times 4}$, and $\tilde{D} \in \mathbb{R}^{4}$ are constant matrices given by $\tilde{A} := \frac{ \partial f}{\partial \delta}\Big|_{(\tilde{\delta}, \tilde{a}, \tilde{z})}$, $\tilde{B} := \frac{ \partial f}{\partial a}\Big|_{(\tilde{\delta}, \tilde{a}, \tilde{z})}$, $\tilde{C}= \frac{ \partial f}{\partial z}\Big|_{(\tilde{\delta}, \tilde{a}, \tilde{z})}$, and $\tilde{D}:=f(\tilde{\delta}, \tilde{a}, \tilde{z})- \tilde{A}\tilde{\delta} - \tilde{B}\tilde{a} - \tilde{C}\tilde{z}$, respectively. Hence, the linearized system dynamics is given by:
\begin{align}
&z(\tau+1) = \tilde{A}\delta(\tau) + \tilde{B} a(\tau) + \tilde{C} z(\tau) + \tilde{D} \nonumber \\
&\implies \tilde{A}\delta(\tau) + \tilde{B} a(\tau) + \tilde{C} z(\tau) - z(\tau +1) + \tilde{D} = 0. 
\end{align}
The equality constraints based on the  system dynamics over the  $T_p$ planning time-steps can be written as:
\begin{align}
   F(\boldsymbol{\Delta}, \boldsymbol{\alpha}, \boldsymbol{Z}) := A\boldsymbol{\Delta} + B\boldsymbol{\alpha} + C \boldsymbol{Z} + D =0,
\end{align}
where $A \in \mathbb{R}^{4T_p \times T_p},  B \in \mathbb{R}^{4T_p \times T_p},C \in \mathbb{R}^{4T_p \times 4T_p}$, and $D \in \mathbb{R}^{4T_p}$ are constant matrices given by:
\begin{align}\label{eq:linear_equality_coeff}
    A &= \begin{bmatrix}
    \tilde{A} & \boldsymbol{0} & \boldsymbol{0} & \cdots \\
    \boldsymbol{0} & \tilde{A} & \boldsymbol{0} & \cdots \\
    \vdots & \vdots & \vdots & \cdots
    \end{bmatrix},   \   B = \begin{bmatrix}
    \tilde{B} & \boldsymbol{0} & \boldsymbol{0} & \cdots \\
    \boldsymbol{0} & \tilde{B} & \boldsymbol{0} & \cdots \\
    \vdots & \vdots & \vdots & \cdots
    \end{bmatrix}, \nonumber \\
        C &= \begin{bmatrix}
    -\boldsymbol{I} & \boldsymbol{0} & \boldsymbol{0} & \cdots \\
    \tilde{C} & -\boldsymbol{I} & \boldsymbol{0} & \cdots \\
     \boldsymbol{0} & \tilde{C} & -\boldsymbol{I} & \cdots \\
    \vdots & \vdots & \vdots & \cdots
    \end{bmatrix},   \   D = \begin{bmatrix}
    \tilde{D} - \tilde{C}z(0) \\
    \tilde{D} \\
    \tilde{D} \\
    \vdots 
    \end{bmatrix},
\end{align}
$\boldsymbol{0}$ and $\boldsymbol{I}$ denote the zero and identity matrix, respectively.
\begin{remark}
To simplify the optimization, we linearly approximate the system dynamics before solving the MPC. This is possible because the control inputs obtained through the MPC are only applied for a single time-step, using a receding horizon control approach~\cite{berberich2022linear}. As a result, any linearization errors from previous time-steps do not affect the MPC optimization.
% We linearly approximate the system dynamics before solving the MPC to simplify the optimization. This is possible because the control inputs obtained through MPC are only applied for one time-step (receding horizon control)~\cite{berberich2022linear}. Therefore, any linearization errors from previous time steps are not introduced in the MPC optimization.  
\end{remark}
% The control inputs are: front steering angle $\delta$ and acceleration $a$. We use Euler discretization to obtain a discrete-time dynamical model in the form: 
% \begin{equation}\label{eq:system_dynamics}
%     z(t+1) = f(z(t),u(t)),
% \end{equation}
% where $z(t) = \begin{bmatrix}x(t)&y(t)&\psi(t)&v(t)\end{bmatrix}^\top$ and $u(t) = \begin{bmatrix}\delta(t)&a(t)\end{bmatrix}^\top$.

% \begin{figure}[H]
%     \centering
%     \includegraphics[width=0.5\textwidth]{images/kinematics_bicycle_dynamics.png}
%     \caption{Kinematic Bicycle Model}
%     \label{fig:kinematic_bicycle_model}
% \end{figure}

% =================================================
% MPC formulation
% =================================================

\subsection{Safety Constraints}\label{sec:control_constraint}

The safety constraints for collision avoidance depend on the  nearby vehicles' trajectory prediction and the vehicle shape model. Let $\mathcal{V}$ denote the set of nearby vehicles surrounding the ego vehicle. Let $\phi(\tau)$ be a trained neural network that jointly predicts the future trajectories of the ego vehicle and its surrounding vehicles for $T_{pred}$ time-steps into the future based on their trajectories for $T_{obs}$ time-steps in the past. 
% These predicted coordinates for the neighboring vehicles are incorporated in the safety constraints ($x_i$ and $y_i$ in \eqref{single_circle_safety}).
% To formalize, we denote the trained neural network as a function $\phi(\tau)$ that maps the observed trajectories to predicted trajectories:
$\phi(\tau)$ is given by:
\begin{align*} %\label{eq:neural_network}
    \phi(\tau) : 
    % \;:\; \nonumber \\
    % &
    &\begin{bmatrix}
    (x(\tau),y(\tau))&\cdots&(x_{N}(\tau),y_{N}(\tau))\\
    \vdots&\vdots&\vdots\\
    \begin{matrix}(x(\tau-T_{obs}+1),\\ \ y(\tau-T_{obs}+1))\end{matrix} &\cdots&\begin{matrix}(x_{N}(\tau-T_{obs}+1),\\ \ y_{N}(\tau-T_{obs}+1))\end{matrix}
    \end{bmatrix}\nonumber\\
    & \hspace{4cm} \downarrow \nonumber \\
    % &&\vspace{3mm}\hspace{-cm}\downarrow\vspace{3mm}\nonumber\\
    &\!\!\!\!\!\! \begin{bmatrix}
    (\hat{x}(\tau+1),\hat{y}(\tau+1))& \cdots&(\hat{x}_{N}(\tau+1),\hat{y}_{N}(\tau+1))
    \end{bmatrix},
\end{align*}
\normalsize
\noindent 
with $T_{pred} = 1$, where the first column represents the positions of the ego vehicle followed by the positions of $N$ surrounding vehicles. Given the buffer of $T_{obs}$ past observations until time-step $\tau$, the coordinates of vehicle $i \in \mathcal{V}$ at time-step $\tau+1$ are represented as:
\begin{align}
    \hat{x}_{i}(\tau+1) = \phi_{i,x}(\tau), \ \qquad \hat{y}_{i}(\tau+1) = \phi_{i,y}(\tau).
\end{align}
Some examples of the neural network $\phi(\tau)$ include social generative adversarial network (SGAN)~\cite{gupta2018social} and graph-based
spatial-temporal convolutional network (GSTCN)~\cite{sheng2022graph}.

\begin{remark}
    Interactive predictions over planning horizon $T_p$ are computed recursively using $\phi(t)$ with $T_{pred} =1$ based on the latest reactive predictions and ego vehicle positions from the MPC's  candidate solution trajectory.
\end{remark}
We model the vehicle shape using a single circle to obtain a smooth and continuously differentiable distance measure to enable gradient-based optimization methods. Let $(x,y)$ and $(\hat{x}_i,\hat{y}_i)$ be the position of the ego vehicle and the predicted positions of the surrounding vehicles  $i \in \mathcal{V}$ (obtained using $\phi(\tau)$), respectively. Let 
$r, r_i \in \mathbb{R}_{>0}$ be the radius of circles modeling ego vehicle and vehicle $i$, respectively. The safety constraint for the ego vehicle w.r.t vehicle $i$ then reads:
% The safety constraints for collision avoidance depend on the vehicle shape model and trajectory prediction of the nearby vehicles which we describe next. 
% \subsubsection{Vehicle Shape Model}
% We model the vehicle shape using a single circle to obtain a smooth and continuously differentiable distance measure to enable gradient-based optimization methods. Let $\mathcal{V}$ denote the set of nearby vehicles surrounding the ego vehicle. Let $(x,y)$ and $(x_i,y_i)$ be the Cartesian coordinates of the center of the ego vehicle and surrounding vehicle $i \in \mathcal{V}$, respectively. The safety constraint for the ego vehicle w.r.t vehicle $i$ is given by:
\begin{align}\label{single_circle_safety}
    d_i(x,y, \hat{x}_i,\hat{y}_i) &= {(x-\hat{x}_i)^2 + (y-\hat{y}_i)^2} \nonumber \\ 
    & \quad \quad \quad \quad \quad - (r+r_i+ \epsilon)^2 >0,
\end{align}
where $\epsilon \in \mathbb{R}_{>0}$ is a safety bound.
\begin{remark}
% A caveat of using a single circle model is that the safety constraints can be conservative, and consequently the feasible solutions could be restrictive in some situations. However, considering the imperfect predictions of the neural network, conservative safety constraints can arguably be acceptable to prioritize vehicle safety. 
Using the single circle model, the safety constraints can be conservative, and consequently, the feasible solutions could be restrictive in some situations. We use it for its simplicity and to reduce the number of safety constraints. Some other alternatives for modeling the vehicle shape include the ellipsoid model~\cite{rosolia2016autonomous} and three circle model~\cite{bae2020cooperation}. 
\end{remark}

\subsection{Formulation of the Optimization problem}
We now present the complete optimization problem for the receding horizon control in a compact form: 
\begin{align}
    \min_{\boldsymbol{\Delta},\boldsymbol{\alpha}, \boldsymbol{Z}}\;\; J=& \Phi_1(\boldsymbol{\Delta}) + \Phi_2(\boldsymbol{\alpha})+ \Phi_3(\boldsymbol{Z}),\label{eq:compact_obj}\\
    \text{subject to  } \;\;& 
    % F(\boldsymbol{\Delta}, \boldsymbol{\alpha}, \boldsymbol{Z})=0\label{eq:compact_eq}\\
    F(\boldsymbol{\Delta}, \boldsymbol{\alpha}, \boldsymbol{Z})  =0, \ \ \ b_i(\boldsymbol{Z})>0, \ i \in \mathcal{V},
    \\
                            & \boldsymbol{\Delta} \in\mathcal{D}, \boldsymbol{\alpha} \in \mathcal{A}, \boldsymbol{Z}\in\mathcal{Z},\label{eq:compact_ineq} \ \text{where} \\
  \  \Phi_1(\boldsymbol{\Delta}) =  \sum_{\tau=0}^{T_p-1}&\lambda_{\delta}\|\delta(\tau)\|^2 + \sum_{\tau=0}^{T_p-1}\lambda_{\Delta\delta}\|\delta(\tau)-\delta(\tau-1)\|^2, \nonumber \\
  \    \Phi_2(\boldsymbol{\alpha}) = \sum_{\tau=0}^{T_p-1} & \lambda_{a}\|a(\tau)\|^2+\sum_{\tau=0}^{T_p-1}\lambda_{\Delta a}\|a(\tau)-a(\tau-1)\|^2, \nonumber \\
 \     \Phi_3(\boldsymbol{Z}) = \sum_{\tau=1}^{T_p} & \lambda_{div}\|y(\tau)-y^{\text{ref}}\|^2 + \sum_{\tau=1}^{T_p}\lambda_{v}\|v(\tau)-v^{\text{ref}}\|^2, \nonumber \\
 & \!\!\!\!\!\!\!\!\!\!\!\!\!\!\!\!\!\!\!\!\!\!\!\!\!\!\!\!\!\!\!\!\!\!\!\!
 b_i(\boldsymbol{Z}) = \begin{bmatrix}
     d_i(x(1), y(1),  \phi_{i,x}(0), \phi_{i,y}(0)) \\
     \vdots \\
     d_i(x(T_p), y(T_p), \phi_{i,x}(T_p-1), \phi_{i,y}(T_p-1)) \\
     \end{bmatrix}. \nonumber
%   \mathcal{D} &= [\delta_{min}, \delta_{max}] : \textit{domain of $\boldsymbol{\Delta}$}, \mathcal{D} \subset \mathbb{R}^{T_p},\\
%     \mathcal{A} &=[a_{min}, a_{max}] : \textit{domain of $\boldsymbol{\alpha}$}, \mathcal{A} \subset \mathbb{R}^{T_p}, \\
%      \mathcal{Z} &= [\boldsymbol{z}_{min}, \boldsymbol{z}_{max}]: \textit{domain of $\boldsymbol{Z}$} , \mathcal{Z} \subset \mathbb{R}^{4T}.
\end{align}
In the next section, we solve the optimization using ADMM to determine a safe and interactive ego vehicle's trajectory.

\section{Solving MPC with ADMM}\label{Sec:ADMM-formulation}
There are many mathematical challenges associated with the MPC problem in Section~\ref{Sec:Mathematical-formulation}. Namely, it has the non-linear system dynamics, non-convex safety constraints, and dependence of the neural network predictions on its predictions in previous time steps ($T_{obs} \neq 1$). We now detail the systematic steps to solve the complex problem using ADMM, addressing the aforementioned mathematical challenges. 
\begin{comment}
In particular, we decompose it into sub-problems, linearize the system dynamics, and utilize non-convex optimization methods. We now solve the optimization problem using the ADMM algorithm.
\end{comment} 

First, we construct a Lagrangian by moving the safety constraints, $b_i(\boldsymbol{Z})>0, \ i \in \mathcal{V}$, in the optimization objective:
\begin{align}
  \!\!\! \min_{\boldsymbol{\Delta},\boldsymbol{\alpha}, \boldsymbol{Z}}J=& \Phi_1(\boldsymbol{\Delta}) + \Phi_2(\boldsymbol{\alpha})+ \Phi_3(\boldsymbol{Z}) - \sum_{i=1}^{N} \lambda_s^Tb_i(\boldsymbol{Z}), \label{eq:compact_obj_final} \\
    &\text{subject to  } \;\ F(\boldsymbol{\Delta}, \boldsymbol{\alpha}, \boldsymbol{Z}) = 0,\label{eq:compact_eq_final}
    \\
                            & \boldsymbol{\Delta} \in\mathcal{D}, \boldsymbol{\alpha} \in \mathcal{A}, \boldsymbol{Z}\in\mathcal{Z},\label{eq:compact_ineq_final}
\end{align}
where $\lambda_s \in \mathbb{R}^{T_p}_{>0}$ is the vector of Lagrange multipliers. 
\begin{remark}
For theoretical analysis, we incorporate safety constraints into the optimization objective, but for our simulation study, we enforce them as hard constraints.
\end{remark}
The optimization problem  \eqref{eq:compact_obj_final}-\eqref{eq:compact_ineq_final} is separable and the optimization variables $\boldsymbol{\Delta},\boldsymbol{\alpha}, \boldsymbol{Z}$ are decoupled in the objective function. %Therefore, we apply ADMM
% \footnote{Note that the Bellman structure cannot be applied because the gradient of the safety constraint must be evaluated with respect to controls of a few steps behind the current step, not just the current step.}
%\cite{boyd2004convex} to find the optimal solution. 
Following the convention in \cite{boyd2004convex}, the augmented Lagrangian is given by:
\vspace{-0.2in}
\begin{multline}
    \mathcal{L}_\rho(\boldsymbol{\Delta},\boldsymbol{\alpha}, \boldsymbol{Z}) = \Phi_1(\boldsymbol{\Delta}) + \Phi_2(\boldsymbol{\alpha}) + \Phi_3(\boldsymbol{Z}) - \sum_{i=1}^{N} \lambda_s^Tb_i(\boldsymbol{Z}) + \\
    \mu^\top F(\boldsymbol{\Delta},\boldsymbol{\alpha}, \boldsymbol{Z})+\left(\frac{\rho}{2}\right)\|F(\boldsymbol{\Delta},\boldsymbol{\alpha}, \boldsymbol{Z})\|^2,
\end{multline}
where $\rho > 0$ is the ADMM Lagrangian parameter and $\mu$ is the dual variable associated with the constraint~\eqref{eq:compact_eq_final}.
The complete algorithm is given by the Algorithm~\ref{alg:mpc_admm}. 
\begin{algorithm}[ht]
    \SetKwInOut{Input}{Input}
    \SetKwInOut{Output}{Output}
    \SetKwInOut{Init}{Init}
    \Init{states $z = z_0$,   controls $\delta = \delta_0, a=a_0$\\
    Surrounding vehicles' position: \\ $(x_i,y_i) = (x_{i,0},y_{i,0})$ for all $i\in \mathcal{V}$}
    \While{$x < x^{\text{ref}}$ and $y \neq y^{\text{ref}}$}{
        Find the optimal control that minimizes the cumulative cost over horizon $T_p$\\
        \Init{$\boldsymbol{\hat{\Delta}} = \boldsymbol{\Delta}_0, \boldsymbol{\hat{\alpha}} = \boldsymbol{\alpha}_0, \boldsymbol{\hat{Z}} = \boldsymbol{Z}_0, \hat{\mu} = \mu_0$}
        \While{convergence criterion is not met}{
            $\boldsymbol{\hat{\Delta}} \leftarrow \argmin_{\boldsymbol{\Delta}} \mathcal{L}_\rho(\boldsymbol{\Delta},\boldsymbol{\hat{\alpha}}, \boldsymbol{\hat{Z}})$ \\
            $\boldsymbol{\hat{\alpha}} \leftarrow \argmin_{\boldsymbol{\alpha}} \mathcal{L}_\rho(\boldsymbol{\hat{\Delta}},\boldsymbol{\alpha}, \boldsymbol{\hat{Z}})$ \\
            $\hat{\boldsymbol{Z}} \leftarrow \argmin_{\boldsymbol{Z}} \mathcal{L}_\rho(\boldsymbol{\hat{\Delta}},\boldsymbol{\hat{\alpha}}, \boldsymbol{Z})$ \\
            $\hat{\mu}\leftarrow \hat{\mu}+\rho(F(\boldsymbol{\hat{\Delta}},\boldsymbol{\hat{\alpha}}, \boldsymbol{\hat{Z}}))$\\
        }
        %\vspace{3mm}
        
        Update the states through non-linear state dynamics 
        % \eqref{eq:non-linear dynamics}
        with first elements of controls \\
        $z \leftarrow f([\boldsymbol{\hat{\Delta}}]_0,[\boldsymbol{\hat{\alpha}}]_0, z)$\\
        %\vspace{3mm}
        
        Observe positions of other vehicles at the current time $t$\\
        $(x_i,y_i) \leftarrow (x_i(t),y_i(t))$ for all $i$
    }
    \caption{MPC with ADMM}\label{alg:mpc_admm}
\end{algorithm}
 Next, we provide details for solving each of the local optimization problems at iteration $k$,
% % \eqref{eq:update_delta}-\eqref{eq:update_z}
for solving the MPC. 

% Note that while the objective functions for the sub-problems \eqref{eq:update_delta}-\eqref{eq:update_a} are convex, the objective function in \eqref{eq:update_z} is non-convex.
\subsection{Update  \texorpdfstring{$\boldsymbol{\Delta}^{(k+1)} = \argmin_{\boldsymbol{\Delta}\in\mathcal{D}}\;\mathcal{L}_{\rho}(\boldsymbol{\Delta},\boldsymbol{\alpha}^{(k)}, \boldsymbol{Z}^{(k)})$}{deltak}}\label{sec:update_delta}
The sub-optimization problem for $\boldsymbol{\Delta}^{(k+1)}$
% \eqref{eq:update_delta}
is given by
\begin{align}
    &\argmin_{\boldsymbol{\Delta}} \ \sum_{\tau=0}^{T_p-1}\lambda_{\delta}\|\delta(\tau)\|^2 + \sum_{\tau=0}^{T_p-1}\lambda_{\Delta\delta}\|\delta(\tau)-\delta(\tau-1)\|^2 \nonumber\\ & \quad \quad \quad \quad\quad\quad +\mu^{k\top}A\boldsymbol{\Delta}+\left(\frac{\rho}{2}\right)\|A\boldsymbol{\Delta} -c_{\boldsymbol{\Delta}}^{(k)}\|^2 , \\
    &\text{subject to}\;\;\;\delta(\tau) \in [\delta_{\min},\delta_{\max}], \nonumber
\end{align}
% Similarly, we linearly approximate $F(\boldsymbol{Z}^{(k+1)},\boldsymbol{\Delta},\boldsymbol{\alpha}^{(k)})$ around $(\boldsymbol{Z}^{(k+1)},\boldsymbol{\Delta}^{(k)},\boldsymbol{\alpha}^{(k)})$, which yields \begin{align}
%     F(\boldsymbol{Z}^{(k+1)},\boldsymbol{\Delta},\boldsymbol{\alpha}^{(k)})&\approx F(\boldsymbol{Z}^{(k+1)},\boldsymbol{\Delta}^{(k)},\boldsymbol{\alpha}^{(k)})+ \nonumber \\ &\partial_{\boldsymbol{\Delta}} F(\boldsymbol{Z}^{(k+1)},\boldsymbol{\Delta}^{(k)},\boldsymbol{\alpha}^{(k)})(\boldsymbol{\Delta}-\boldsymbol{\Delta}^{(k)}).
% \end{align} 
% The problem now reads
% \begin{align}
%     \argmin_{\boldsymbol{\Delta}} &\sum_{\tau=0}^{T_p-1}\lambda_{\delta}\|\delta(\tau)\|^2 + \sum_{\tau=0}^{T_p-1}\lambda_{\Delta\delta}\|\delta(\tau)-\delta(\tau-1)\|^2+ \nonumber \\
%     &\mu^{k\top}(A_{\boldsymbol{\Delta}}^{(k)}\boldsymbol{\Delta})+\left(\frac{\rho}{2}\right)\|A_{\boldsymbol{\Delta}}^{(k)} \boldsymbol{\Delta} -c_{\boldsymbol{\Delta}}^{(k)}\|^2_2\\
%     \text{subject to}\;\;\;&\delta(\tau) \in [\delta_{\min},\delta_{\max}]
% \end{align}∥
where $ c_{\boldsymbol{\Delta}}^{(k)} = A\boldsymbol{\Delta}^{(k)}-F(\boldsymbol{\Delta}^{(k)},\boldsymbol{\alpha}^{(k)}, \boldsymbol{Z}^{(k)})$. It is a convex problem; hence, we can use a canonical convex optimization algorithm~\cite{boyd2004convex} to find the optimal solution.

%% update a
\subsection{Update \texorpdfstring{$\boldsymbol{\alpha}^{(k+1)}=\argmin_{\boldsymbol{\alpha}\in\mathcal{A}}\;\mathcal{L}_{\rho}(\boldsymbol{\Delta}^{(k+1)},\boldsymbol{\alpha}, \boldsymbol{Z}^{(k)})$}{ak}} \label{sec:update_a}
The sub-optimization problem for $\boldsymbol{\alpha}^{(k+1)}$ is given by
\begin{align}
    &\argmin_{\boldsymbol{\alpha}}\sum_{\tau=0}^{T_p-1}\lambda_{a}\|a(\tau)\|^2+\sum_{\tau=0}^{T_p-1}\lambda_{\Delta a}\|a(\tau)-a(\tau-1)\|^2 \nonumber \\
    & \quad  \quad  \quad \quad \quad \quad + \mu^{k\top}B \boldsymbol{\alpha}+\left(\frac{\rho}{2}\right)\|B\boldsymbol{\alpha} - c_{\boldsymbol{\alpha}}^{(k)}\|^2,\\
    &\text{subject to}\;\;\;a(\tau) \in [a_{\min},a_{\max}], \nonumber
\end{align}
% Similarly, we linearly approximate $F(\boldsymbol{Z}^{(k+1)},\boldsymbol{\Delta}^{(k+1)},\boldsymbol{\alpha})$ around $(\boldsymbol{Z}^{(k+1)},\boldsymbol{\Delta}^{(k+1)},\boldsymbol{\alpha}^{(k)})$, which yields 
% \begin{align}
%     F(\boldsymbol{Z}^{(k+1)},\boldsymbol{\Delta}^{(k+1)},\boldsymbol{\alpha}) &\approx F(\boldsymbol{Z}^{(k+1)},\boldsymbol{\Delta}^{(k+1)},\boldsymbol{\alpha}^{(k)})+ \nonumber \\
%     &\partial_{\boldsymbol{\alpha}} F(\boldsymbol{Z}^{(k+1)},\boldsymbol{\Delta}^{(k+1)},\boldsymbol{\alpha}^{(k)})(\boldsymbol{\alpha}-\boldsymbol{\alpha}^{(k)}).
% \end{align} 
where $ c_{\boldsymbol{\alpha}}^{(k)} = B\boldsymbol{\alpha}^{(k)}-F(\boldsymbol{\Delta}^{(k+1)},\boldsymbol{\alpha}^{(k)}, \boldsymbol{Z}^{(k)})$. It is a convex problem; hence, we can use a canonical convex optimization algorithm~\cite{boyd2004convex} to find the optimal solution.

% The problem now reads
% \begin{align}
%     \argmin_{\boldsymbol{\alpha}} &\sum_{\tau=0}^{T_p-1}\lambda_{a}\|a(\tau)\|^2+\sum_{\tau=0}^{T_p-1}\lambda_{\Delta a}\|a(\tau)-a(\tau-1)\|^2+ \nonumber \\
%     &\mu^{k\top}(A_{\boldsymbol{\alpha}}^{(k)} \boldsymbol{\alpha})+\left(\frac{\rho}{2}\right)\|A_{\boldsymbol{\alpha}}^{(k)}\boldsymbol{\alpha}-c_{\boldsymbol{\alpha}}^{(k)}\|^2_2\\
%     \text{subject to}\;\;\;&a(\tau) \in [a_{\min},a_{\max}]
% \end{align}

% where
% \begin{align}
%     A_{\boldsymbol{\alpha}}^{(k)} &= \partial_{\boldsymbol{\alpha}} F(\boldsymbol{Z}^{(k+1)},\boldsymbol{\Delta}^{(k+1)},\boldsymbol{\alpha}^{(k)})\\ 
%     c_{\boldsymbol{\alpha}}^{(k)} &= A_{\boldsymbol{\alpha}}^{(k)}\boldsymbol{\alpha}^{(k)}-F(\boldsymbol{Z}^{(k+1)},\boldsymbol{\Delta}^{(k+1)},\boldsymbol{\alpha}^{(k)})
% \end{align}
% The problem is now a convex problem and we can use a canonical convex optimization algorithm, e.g., gradient descent, to find the optimal solution.

\subsection{Update \texorpdfstring{$\boldsymbol{Z}^{(k+1)}=\argmin_{\boldsymbol{Z}\in\mathcal{Z}}\;\mathcal{L}_{\rho}(\boldsymbol{\Delta}^{(k+1)},\boldsymbol{\alpha}^{(k+1)}, \boldsymbol{Z})$}{zk}} \label{sec:update_z}

The sub-optimization problem for $\boldsymbol{Z}^{(k+1)}$ is given by 
% \eqref{eq:update_z} can be written as:
\begin{align}
    &\argmin_{\boldsymbol{Z}} \sum_{\tau=1}^{T_p}\lambda_{div}\|y(\tau)-y^{\text{ref}}\|^2+ \sum_{\tau=1}^{T_p}\lambda_{v}\|v(\tau)-v^{\text{ref}}\|^2 \nonumber \\
    &  \quad - \sum_{i=1}^{N} \lambda_s^Tb_i(\boldsymbol{Z}) +  \mu^{k\top}C\boldsymbol{Z}+\left(\frac{\rho}{2}\right)\|C\boldsymbol{Z}-c_{\boldsymbol{Z}}^{(k)}\|^2, \label{eq:update_z_non_convex} \\
    &\text{subject to } z(\tau) \in [z_{\min}, z_{\max}], \label{eq:update_z_non_convex_1}
    % g_i(z(\tau+1)) > 0, \forall i\in\{1,\cdots,N\}, \ \tau \in \{0, \ldots, T_p-1\}, \\
    % & \red{x(\tau)<x(\tau+1), \ \ \tau \in \{0, \ldots, T_p-1\}},
\end{align}
where $c_{\boldsymbol{Z}}^{(k)} = C\boldsymbol{Z}^{(k)}-F(\boldsymbol{\Delta}^{(k+1)},\boldsymbol{\alpha}^{(k+1)}, \boldsymbol{Z}^{(k)})$. Due to the nonconvexity of the neural network in $b_i(\boldsymbol{Z})$, the objective function~\eqref{eq:update_z_non_convex} is non-convex.
% Therefore, we can apply sequential quadratic programming (SQP)~\cite{boggs2000sequential} to find the optimal solution. However,  solving an SQP involves computing the gradient and hessian of the neural network $\phi(\cdot)$ w.r.t the states of the ego vehicle $\boldsymbol{Z}$. Furthermore, since we have to recursively predict the trajectory of the vehicles for $T_p$ time steps, the network predictions depend on the older predictions in the previous time steps. This makes the gradient computation expensive for longer planning horizons due to the increasing complexity of the neural network's computational graph. Therefore, Quasi-Newton methods are preferred to solve the optimization to avoid expensive Hessian computation at each step. 
We prefer the Quasi-Newton method for optimization to avoid expensive Hessian computation at each step. Hence, we utilize BFGS-SQP method~\cite{curtis2017bfgs}, which employs BFGS Hessian approximations within a sequential quadratic optimization, and does not assume any special structure in the objective or constraints. For a solver, we use PyGranso~\cite{liang2021ncvx}, a PyTorch-enabled port of GRANSO, that enables gradients computation by back-propagating the neural network's gradients at each iteration. 
% incorporating auto-differentiation, GPU acceleration, and tensor input. PyGranso's auto-differentiation allows us to efficiently compute the gradients by efficiently back-propagating gradients through the neural network at each iteration. 
% \di{wasn't this the bottleneck that makes us slower than realtime? if so, maybe we shouldn't call it efficient...}

% Note that updating the control input trajectories $\boldsymbol{\Delta}$ and $\boldsymbol{\alpha}$ are canonical convex optimization problems and existing solvers can find solutions very efficiently. However, updating the state trajectory $\boldsymbol{Z}$ has a larger complexity in the problem due to the presence of the non-convex neural network predictions. Therefore, computing the gradient of the objective function requires back-propagation of the gradients through the neural network at each iteration. 

\begin{remark}
The state trajectory $\boldsymbol{Z}$ update has a larger complexity in the problem due to the presence of the non-convex neural network predictions. To expedite the $\boldsymbol{Z}$ update, an offline-trained function approximator such as a neural network can be utilized to estimate the gradients of the original neural network. The training dataset for gradient approximator can be generated using automatic differentiation or central differences approximations with original network.
% To make $\boldsymbol{Z}$ update faster, a neural network may be trained offline to estimates the gradients of the original neural network. This can be done by discretizing the feasible space around the ego-vehicle with the ego-vehicle at the center of the coordinate frame. 
\end{remark}

Henceforth, we refer to our method as ADMM-NNMPC.

\section{Convergence of MPC with ADMM}\label{Convergence of ADMM}
% In this section, we study the convergence of Algorithm~\ref{alg:mpc_admm} for solving the MPC optimization that incorporates the non-convex neural network predictions.
Due to the inherent non-convexity of the neural network, the rigorous convergence analysis of ADMM in \cite{boyd2011distributed} is not readily applicable.
% \bae{I saw this sentence somewhat redundant. We can start with something like: With the non-convexity inherited in NN, the rigorous convergence analysis of ADMM [cite] is not readily applicable.}
Thus, we extend the convergence analysis of ADMM with an integrated neural network, i.e., the convergence of the inner while loop in Algorithm~\ref{alg:mpc_admm}. We first make the following assumptions on the neural network:% that is trained to predict the joint interactive motion of the nearby vehicles.

\begin{enumerate}
    \item [(A1)] At any time-step $\tau \in [0, T_p]$, the neural network's outputs are bounded, i.e. $|\phi_{i,x}(\tau)| \le s_x$ and $|\phi_{i,y}(\tau)| \le s_y$, $i \in \mathcal{V}$, where $s_x, s_y \in \mathbb{R}_{>0}$ are constants.
    
     \item [(A2)]  At any time-step $\tau \in [0, T_p]$, the gradients of the neural network's outputs w.r.t the input ego trajectory exist and are bounded, i.e. $\|\frac{\partial{\phi_{i,x}}(t)}{\partial \boldsymbol{Z}}\|_{\infty} \le\theta_x$ and $\|\frac{\partial{\phi_{i,y}(t)}}{\partial \boldsymbol{Z}}\|_{\infty} \le \theta_y$ for all $i \in \mathcal{V}$, where $\theta_x, \theta_y \in \mathbb{R}_{>0}$ are constants and $\|\cdot \|_{\infty}$ is the max. norm of a vector.
     
    % \item [(A3)] The outputs of the neural network are Lipschitz continuous, i.e. $||\phi_{i,x}(\boldsymbol{Z}_1) - \phi_{i,x}(\boldsymbol{Z}) || \le
    % L_{\phi}||\boldsymbol{Z}_1- \boldsymbol{Z}_2||$ and $||\phi_{i,y}(\boldsymbol{Z}_1) - \phi_{i,y}(\boldsymbol{Z}_2) || \le
    % L_{\phi}||\boldsymbol{Z}_1- \boldsymbol{Z}_2||$ for all $i \in \mathcal{V}$, $t \in \mathcal{T_p}$, where $L_{\phi} \in \mathbb{R}_{>0}$ is the Lipschitz constant.
    
    \item [(A3)] At any time-step $\tau \in [0, T_p]$, the neural network's outputs are Lipschitz differentiable, i.e. $||\nabla\phi_{i,x}(\boldsymbol{Z}_1) - \nabla \phi_{i,x}(\boldsymbol{Z}_2) || \le
    L_{\nabla\phi}||\boldsymbol{Z}_1- \boldsymbol{Z}_2||$ and $||\nabla\phi_{i, y}(\boldsymbol{Z}_1) - \nabla \phi_{i, y}(\boldsymbol{Z}_2) || \le
    L_{\nabla\phi}||\boldsymbol{Z}_1- \boldsymbol{Z}_2||$ for all $i \in \mathcal{V}$, $\boldsymbol{Z}_1, \boldsymbol{Z}_2 \in \mathcal{Z}$,  where $L_{\nabla \phi} \in \mathbb{R}_{>0}$ is the Lipschitz constant for the neural network's gradient. \end{enumerate}

Assumptions (A1)-(A3) are sufficient conditions under which the objective function~\eqref{eq:compact_obj_final} is Lipschitz differentiable, i.e., it is differentiable and its gradient is Lipschitz continuous. This allows us to establish the convergence of Algorithm~\ref{alg:mpc_admm}. Assumption (A1) is satisfied for a trained neural network for a bounded input space. Furthermore, neural network outputs can be clipped based on the feasible region. Lastly, neural networks with $C^2$ activation functions such as Gaussian Error Linear Unit (GELU)~\cite{hendrycks2016gaussian} and Smooth Maximum Unit  (SMU)~\cite{biswas2021smu} satisfy assumptions (A2)-(A3).
% Since a twice differentiable function is Lipschitz differentiable~\cite{nesterov2003introductory}, assumptions (A2) and (A3) can be satisfied by using a neural network architecture with differentiable activation functions such as ELU.  

\begin{remark}
Assumptions (A1)-(A3) are sufficient conditions and not necessary conditions. If the neural network architecture is unknown or it doesn't satisfy the assumptions, knowledge distillation~\cite{mirzadeh2020improved} can be used to train a smaller (student)  network that satisfies the assumptions from the large (teacher) pre-trained network.
% . Specifically, the large (teacher) pre-trained network can be used to train a smaller (student) network that satisfies the assumptions of the neural network.   
\end{remark}

\begin{theorem}{[\textbf{Convergence of MPC with ADMM}]}\label{thm1}
Under the assumptions (A1)–(A3), the inner while loop in Algorithm~\ref{alg:mpc_admm} converges subsequently for any sufficiently large $\rho  > \max \{1, (1 + 2\sigma_{\min}(C))L_{J}M\}$, where $\sigma_{\min}(C)$ is the smallest positive singular value of $C$ in~\eqref{eq:linear_equality_coeff}, $L_J$ is the Lipschitz constant for $J$ in~\eqref{eq:compact_obj_final}, and $M$ is the Lipschitz constant for sub-minimization paths as defined in Lemma~\ref{lemma:Lipschitz sub-minimization paths}. Therefore, starting from any $\boldsymbol{\Delta}^{(0)}, \boldsymbol{\alpha}^{(0)}, \boldsymbol{Z}^{(0)}, \mu^{(0)}$, it generates a sequence that
is bounded, has at least one limit point, and that each limit point $\boldsymbol{\Delta}^*, \boldsymbol{\alpha}^*, \boldsymbol{Z}^*, \mu^*$ is a stationary
point of $\mathcal{L}_{\rho}$ satisfying $\nabla \mathcal{L}_{\rho} (\boldsymbol{\Delta}^*, \boldsymbol{\alpha}^*, \boldsymbol{Z}^*, \mu^*)=0$.
\end{theorem}

We prove Theorem~\ref{thm1} using Lemmas~\ref{lemma: feasibility}-\ref{Lemma: Lipschitz smoothness}.
\begin{lemma}{[\textbf{Feasibility}]}\label{lemma: feasibility}
Let $Q:= [A, B]$. Then $\text{Im}(Q)$ $\subseteq \text{Im}(C)$, where $\text{Im}(\cdot)$ returns the image of a matrix, and $A,B,$ and $C$ is defined in~\eqref{eq:linear_equality_coeff}.
\end{lemma}
\begin{proof}
 See Appendix~\ref{feasibility proof Appendix} for the proof.
\end{proof}

\begin{lemma}{[\textbf{Lipschitz sub-minimization paths}]}\label{lemma:Lipschitz sub-minimization paths}
The following statements hold for the optimization problem:
\begin{enumerate}
    \item[(i)] For any fixed $\boldsymbol{\alpha}, \boldsymbol{Z}$, 
    % $\argmin_{\boldsymbol{\Delta}} \{J(\boldsymbol{\Delta}, \boldsymbol{\alpha}, \boldsymbol{Z})\} : A\boldsymbol{\Delta} = u\}$ has a unique minimizer. 
    $H_1 : Im(A) \rightarrow \mathbb{R}^{T_p}$ defined by $H_1(u) \triangleq \argmin_{\boldsymbol{\Delta}} \{J(\boldsymbol{\Delta}, \boldsymbol{\alpha}, \boldsymbol{Z}) : A\boldsymbol{\Delta} = u\}$ is unique and a Lipschitz continuous map.
    \item[(ii)] For any fixed $\boldsymbol{\Delta}, \boldsymbol{Z}$,
    % $\argmin_{\boldsymbol{\alpha}} \{J(\boldsymbol{\Delta}, \boldsymbol{\alpha}, \boldsymbol{Z})\} : B\boldsymbol{\alpha} = u\}$ has a unique minimizer. 
    $H_2 : Im(B) \rightarrow \mathbb{R}^{T_p}$ defined by $H_2(u) \triangleq \argmin_{\boldsymbol{\alpha}} \{J(\boldsymbol{\Delta}, \boldsymbol{\alpha}, \boldsymbol{Z}) : B\boldsymbol{\alpha} = u\}$ is unique and a Lipschitz continuous map.
    \item[(iii)] For any fixed $\boldsymbol{\Delta}, \boldsymbol{\alpha}$, 
    % $\argmin_{\boldsymbol{Z}} \{J(\boldsymbol{\Delta}, \boldsymbol{\alpha}, \boldsymbol{Z})\} : C\boldsymbol{Z} = u\}$ has a unique minimizer.
    $H_3 : Im(C) \rightarrow \mathbb{R}^{4T_p}$ defined by $H_3(u) \triangleq \argmin_{\boldsymbol{Z}} \{J(\boldsymbol{\Delta}, \boldsymbol{\alpha}, \boldsymbol{Z}) : C\boldsymbol{Z} = u\}$ is unique and a Lipschitz continuous map,
\end{enumerate}
where $A,B,$ and $C$ is defined in~\eqref{eq:linear_equality_coeff}. Moreover, $H_1, H_2, H_3$ have a universal Lipschitz constant $M>0$.
\end{lemma}

% \begin{proof}
% \begin{align}
%   &A_{\boldsymbol{\Delta}}  =  \partial_{\boldsymbol{\Delta}} F( \boldsymbol{\Delta}^{(k)}, \boldsymbol{\alpha}^{(k)}, \boldsymbol{Z}^{(k)}) = \\
%   &\begin{bmatrix}
%     k_1(0) & 0&0& \ldots& 0\\
%   k_2(0) & 0&0& \ldots& 0\\
%     k_3(0) &0& 0& \ldots& 0\\
%     0 & 0& 0&\ldots& 0\\
%     0 &  k_1(0)& 0  & \ldots& 0\\
%     0 & k_2(0) & 0& \ldots& 0\\
%     0 & k_3(0) & 0& \ldots& 0\\
%     0 & 0& 0 &\ldots& 0\\
%     \vdots &  \vdots &   \vdots&  \vdots&  \vdots\\
%     0  & 0 & 0 & \cdots & 0
%     \end{bmatrix}
% \end{align}
% where $k_1(\cdot)=v(\cdot)sin(\psi(\cdot)+\beta(\cdot))\Delta t \frac{\partial \beta(\cdot)}{\partial \delta}$, $k_2(\cdot)= -v(\cdot)cos(\psi(\cdot)+\beta(\cdot))\Delta t \frac{\partial \beta(\cdot)}{\partial \delta}$, and  $k_3(\cdot)=-\frac{v(\cdot)cos(\beta(\cdot))\Delta t}{l_r}\frac{\partial \beta(\cdot)}{\partial \delta}$.
% \begin{equation}
%     A_{\boldsymbol{\alpha}} = \partial_{\boldsymbol{\alpha}} F( \boldsymbol{\Delta}^{(k)}, \boldsymbol{\alpha}^{(k)}, \boldsymbol{Z}^{(k)}) = \begin{bmatrix}
%     0 & 0& \ldots& 0\\
%     0 & 0& \ldots& 0\\
%     0 &0& \ldots& 0\\
%     -\Delta t & 0&\ldots& 0\\
%     0 & 0& \ldots& 0\\
%     0 & 0& \ldots& 0\\
%     0 &0& \ldots& 0\\
%     0 & -\Delta t & \ldots& 0\\
%     \vdots &  \vdots &  \vdots&  \vdots\\
%     0  & 0  & \cdots & -\Delta t
%     \end{bmatrix}
% \end{equation}
\begin{proof}
  See Appendix~\ref{Lipschitz sub_minimization proof Appendix} for the proof.
\end{proof}
% \bae{Consider putting proofs of Lemma 2 and Lemma 3 and proof of theorem 1 in Appendix for consistency}

\begin{lemma}{[\textbf{Lipschitz Differentiability}]}\label{Lemma: Lipschitz smoothness}
Under the assumptions (A1)-(A3), the objective function $J(\boldsymbol{\Delta}, \boldsymbol{\alpha}, \boldsymbol{Z})$ in~\eqref{eq:compact_obj_final} is Lipschitz differentiable.
\end{lemma}

\begin{proof}
 See Appendix~\ref{Lipschitz smoothness proof Appendix} for the proof.
\end{proof}

\textit{Proof of Theorem 1:} 
 See Appendix~\ref{convergence theorem proof appendix} for the proof. $\hfill \blacksquare$

\begin{figure}
    \centering
	\begin{subfigure}[b]{0.11\textwidth}
	    \centering
        \includegraphics[width=1.15\linewidth, height=1.15\linewidth, keepaspectratio]{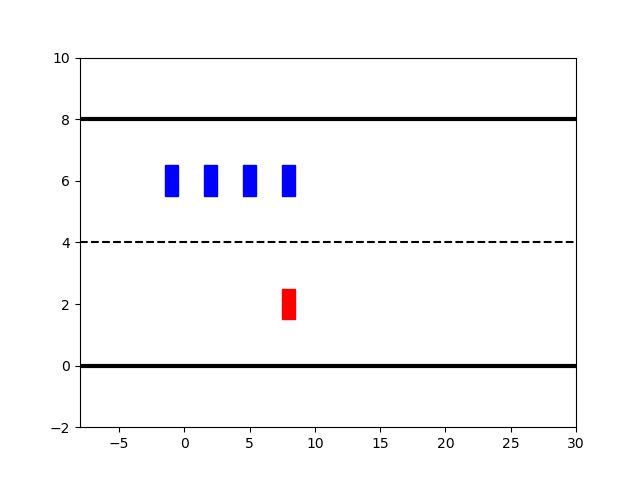}
        \caption{$t=0$}
        \label{fig:two_lanes t=0}
    \end{subfigure}
	\begin{subfigure}[b]{0.11\textwidth}
	    \centering
        \includegraphics[width=1.15\linewidth, height=1.15\linewidth, keepaspectratio]{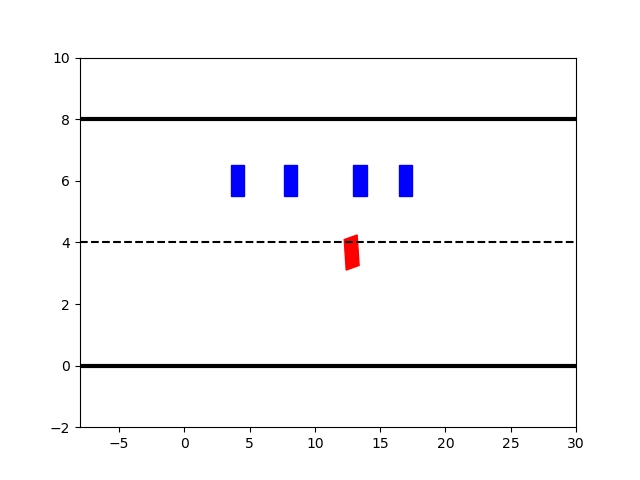}
        \caption{$t=5$}
        \label{fig:two_lanes t=5}
    \end{subfigure}
    \begin{subfigure}[b]{0.11\textwidth}
	    \centering
        \includegraphics[width=1.15\linewidth, height=1.15\linewidth, keepaspectratio]{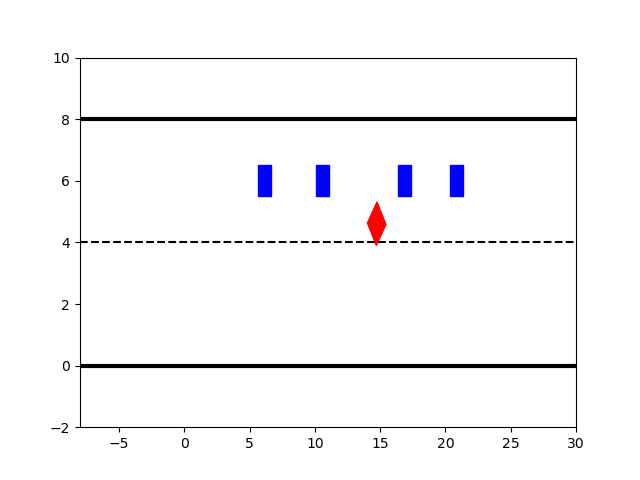}
        \caption{$t=11$}
        \label{fig:two_lanes t=7}
    \end{subfigure}
    \begin{subfigure}[b]{0.11\textwidth}
	    \centering
        \includegraphics[width=1.15\linewidth, height=1.15\linewidth, keepaspectratio]{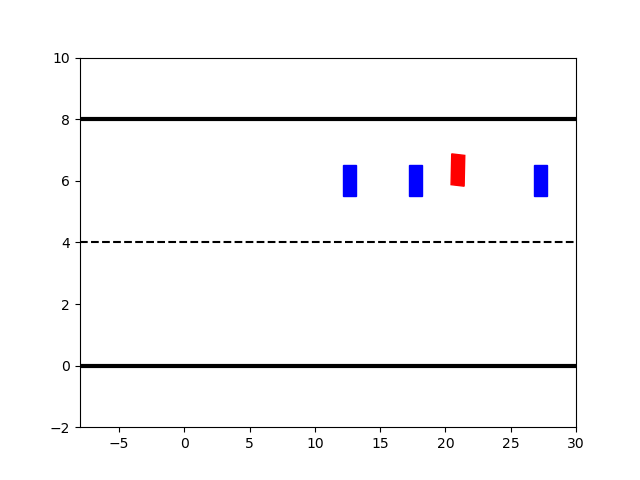}
        \caption{$t=13$}
        \label{fig:two_lanes t=13}
    \end{subfigure}
    \caption{\footnotesize{\textbf{Two lane scenario:} (a)-(d) shows the ADMM-NNMPC solution in a two-lane scenario after $0, 5, 7,$ and $13$ time steps, respectively. 
    % The ego vehicle (red) intends to merge in the left lane occupied by four other vehicles (blue) with a narrow inter-vehicle gap.
    The ego vehicle (red) opens a gap by nudging the vehicles to change their speeds.  
    }}
    \label{fig:ADMM_solution_two_lanes}
\end{figure}
\begin{figure}
    \centering
    \begin{subfigure}[b]{0.11\textwidth}
	    \centering
        \includegraphics[width=1.15\linewidth, height=1.15\linewidth, keepaspectratio]{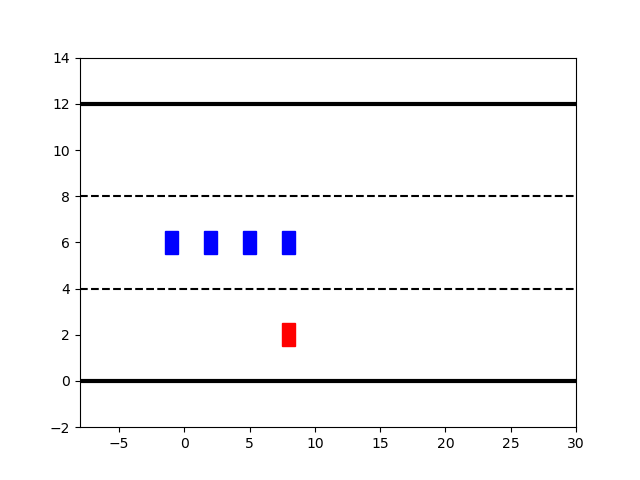}
        \caption{$t=0$}
        \label{fig:three_lanes t=0}
    \end{subfigure}
	\begin{subfigure}[b]{0.11\textwidth}
	    \centering
        \includegraphics[width=1.15\linewidth, height=1.15\linewidth, keepaspectratio]{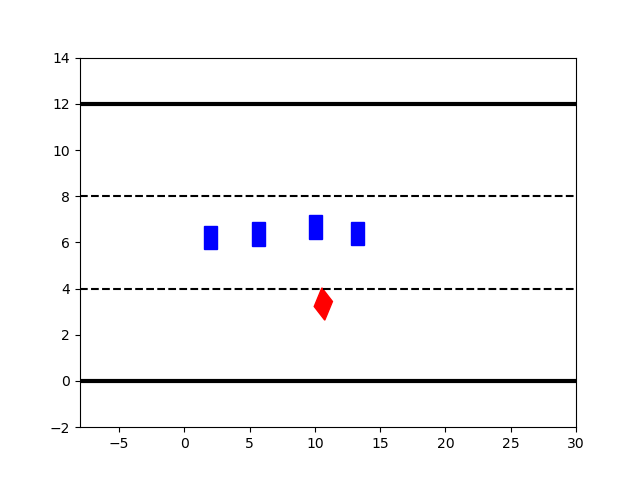}
        \caption{$t=3$}
        \label{fig:three_lanes t=3}
    \end{subfigure}
    \begin{subfigure}[b]{0.11\textwidth}
	    \centering
        \includegraphics[width=1.15\linewidth, height=1.15\linewidth, keepaspectratio]{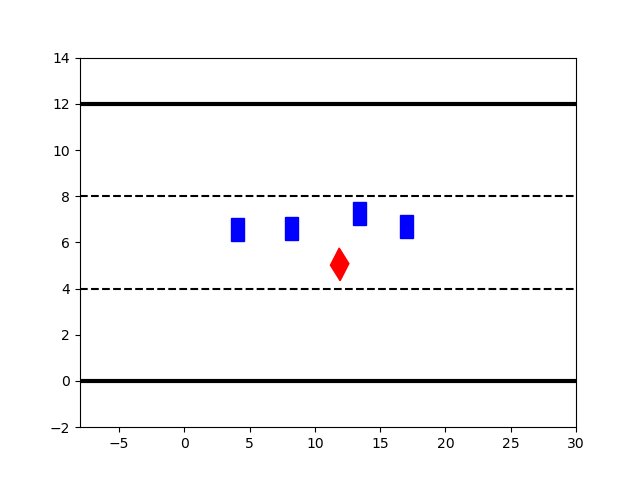}
        \caption{$t=5$}
        \label{fig:three_lanes t=5}
    \end{subfigure}
    \begin{subfigure}[b]{0.11\textwidth}
	    \centering
        \includegraphics[width=1.15\linewidth, height=1.15\linewidth, keepaspectratio]{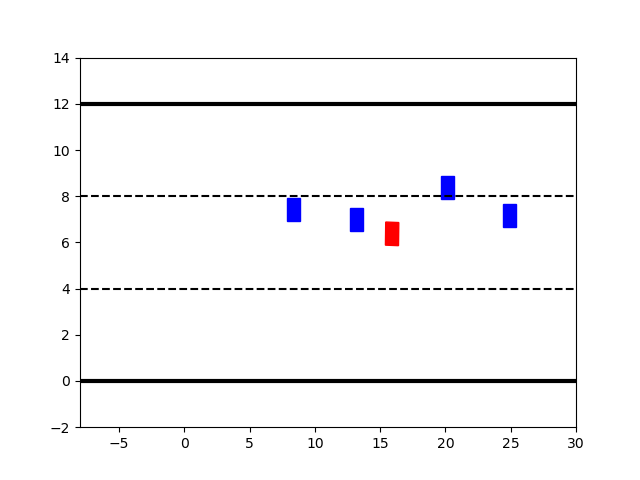}
        \caption{$t=9$}
        \label{fig:three_lanes t=9}
    \end{subfigure}
    \caption{\footnotesize{\textbf{Three lane scenario:} (a)-(d) shows the ADMM-NNMPC solution in a three-lane scenario after $0, 3, 5,$ and $9$ time steps, respectively. 
    % The ego vehicle (red) intends to merge in the left lane occupied by four other vehicles (blue) with a narrow inter-vehicle gap.
    The ego vehicle (red) opens a gap for itself by nudging the vehicles to transition into the left-most lane. 
    }}
    \label{fig:ADMM_solution_three_lanes}
\end{figure}

\begin{table}[]
\centering
\begin{tabular}{|l|l|l|}
\hline
\textbf{Param}                     & \textbf{Description}                            & \textbf{Value} \\ \hline
$\lambda_{div}$           & Weight on divergence from target lane  & 1.0   \\ \hline
$\lambda_{v}$             & Weight on divergence from target speed & 1.0   \\ \hline
$\lambda_{\delta}$        & Weight on steering angle               & 0.6   \\ \hline
$\lambda_{a}$             & Wight on acceleration                  & 0.4   \\ \hline
$\lambda_{\Delta \delta}$ & Weight on steering rate                & 0.4   \\ \hline
$\lambda_{\Delta a}$      & Weight on jerk                         & 0.2   \\ \hline
$\rho$      & ADMM Lagrangian parameter                         & 100   \\ \hline
\end{tabular}
\caption{\label{table:params} \footnotesize Objective function coefficients}
\end{table}
\begin{table}[]
\centering
\resizebox{0.45\textwidth}{!}{%
\begin{tabular}{|c|cc|cc|}
\hline
\multirow{2}{*}{} & \multicolumn{2}{c|}{\textbf{Two-Lane Scenario}} & \multicolumn{2}{c|}{\textbf{Three-Lane Scenario}} \\ \cline{2-5} 
                     & \multicolumn{1}{c|}{NNMPC}         & ADMM-NNMPC & \multicolumn{1}{c|}{NNMPC} & ADMM-NNMPC \\ \hline
{$t_{merge}$} & \multicolumn{1}{c|}{Fails after 17} & 9          & \multicolumn{1}{c|}{29}    & 7          \\ \hline
{$C_{\max}$}  & \multicolumn{1}{c|}{80.9}          & 62         & \multicolumn{1}{c|}{114.6} & 53.3       \\ \hline
{$d_{\min}$}  & \multicolumn{1}{c|}{0.91}          & 2.41       & \multicolumn{1}{c|}{2.97}  & 2.66       \\ \hline
\end{tabular}%
}
\caption{\label{table:results} \footnotesize Simulation results for ADMM-NNMPC and NNMPC in the two-lane and three-lane scenario. $t_{merge}$ are the number of time steps taken by the ego vehicle to merge into the target lane. $C_{\max}$ and $d_{\min}$ are the maximum cost and minimum distance between the ego vehicle and other vehicles at any point of the simulation, respectively.}
\end{table}

\section {Simulation Study}\label{Sec: Numerical Illustrations}
We now present the simulation results for ADMM-NNMPC. Figure~\ref{fig:ADMM_solution_two_lanes} and~\ref{fig:ADMM_solution_three_lanes} show the vehicles' positions at different time steps in two scenarios in which the ego vehicle (red) intends to merge into the left lane which is occupied by four other vehicles (blue) with a narrow inter-vehicle gap. In the two-lane scenario (Fig.~\ref{fig:ADMM_solution_two_lanes}), other vehicles can only change their speeds, while in the three-lane scenario (Fig.~\ref{fig:ADMM_solution_three_lanes}), other vehicles can also move laterally to transition into the leftmost lane. 
% Figs.~\ref{fig:ADMM_solution_two_lanes} and ~\ref{fig:ADMM_solution_three_lanes} show the position of the vehicles at different time steps in the two-lane and three-lane scenarios, respectively. 
The other vehicles' positions at different time steps match the neural network's predictions, and hence, the ego vehicle's actions affect the trajectory of the other vehicles. In both scenarios, the ego vehicle is able to interact with the other agents and open a gap for itself to merge into.

% \begin{figure}
%  \centering
%  \includegraphics[width=0.6\linewidth, height=0.6\linewidth, keepaspectratio]{images/fig_traj.png}
%  \caption{\footnotesize Bird-eye view of the track}
%  \label{fig:route}
% \end{figure}

\begin{figure}
    \centering
	\begin{subfigure}[b]{0.21\textwidth}
	    \centering
        \includegraphics[width=1.1\linewidth, height=1\linewidth, keepaspectratio]{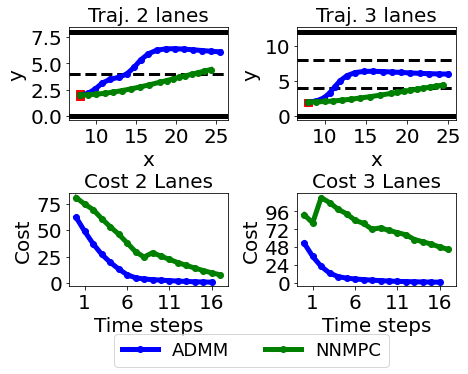}
        \caption{}
        \label{fig:fig_traj}
    \end{subfigure}
    ~~
	\begin{subfigure}[b]{0.21\textwidth}
	    \centering
        \includegraphics[width=1.1\linewidth, height=1\linewidth, keepaspectratio]{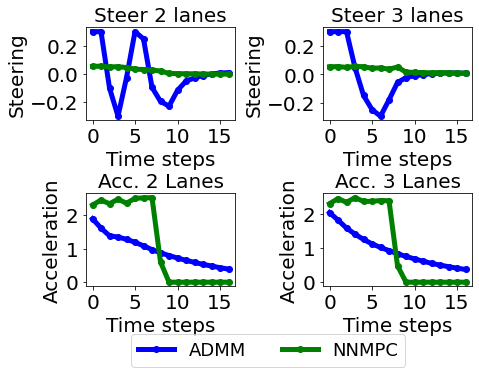}
        \caption{}
        \label{fig:fig_control}
    \end{subfigure}
        \caption{\footnotesize{(a) compares the trajectory (top) and cost (bottom) of the ADMM-NNMPC and NNMPC solutions in the two-lane (left) and three-lane (right) scenarios until $x_{ref} = 25$. (b)  compares the steering (top) and acceleration (bottom) trajectories for ADMM-NNMPC and NNMPC solutions in the two-lane (left) and three-lane (right) scenarios.
        % the trajectories and cost obtained by ADMM-NNMPC and NNMPC at different time steps in (a) two-lane scenario and (b) three-lane scenario.
        % \bae{Barely readable. Make the plot larger, or at least make the legend larger if space concerned.}
        }}
    \label{fig:numerical}
\end{figure}

We compare ADMM-NNMPC with a baseline method called NNMPC~\cite{bae2022lane} on the two-lane and three-lane scenarios by utilizing the same cost function (cost function coefficients listed in Table~\ref{table:params}) and $T_p=8$ time steps. NNMPC generates trajectory candidates by computing a finite set of spiral curves from the source lane to the target lane and selects the candidate with minimum cost. In both methods, we use a trained SGAN neural network~\cite{gupta2018social} for interactive motion prediction of the other vehicles.
% with the baseline NNMPC method on the two-lane and three-lane scenarios by utilizing the same cost function for the two methods (cost function coefficients listed in Table~\ref{table:params}). 
Table~\ref{table:results} compares the simulation results for the baseline NNMPC and ADMM-NNMPC in the two-lane and three-lane scenarios. In the two-lane scenario, while the ADMM-NNMPC successfully merges in the left lane, the NNMPC method fails to make a lane change due to limited trajectory candidates. In the three-lane scenario, ADMM-NNMPC successfully switches lanes much faster than NNMPC. Furthermore, ADMM-NNMPC outperforms NNMPC in terms of maximum cost and minimum distance from other vehicles in both scenarios.%\bae{Weird formatting issue}

Figure~\ref{fig:fig_traj} compares the trajectory (top) and cost (bottom) of the ADMM-NNMPC and NNMPC solutions in the two-lane (left) and three-lane (right) scenarios until $x_{ref} = 25$. In both scenarios, while ADMM-NNMPC successfully merges into the left lane, NNMPC fails to switch lanes before $x_{ref}$ due to limited trajectory candidates.
% Figure~\ref{fig:cost_comparison} compares the cost of the solutions obtained by ADMM and NNMPC at each time step in two scenarios. 
% While NNMPC uses a heuristic method to generate the trajectory candidates, ADMM solves the optimization problem.
Furthermore, the ADMM-NNMPC's cost is lower than the NNMPC solution at every time step since ADMM-NNMPC solves the optimization. Figure~\ref{fig:fig_control} compares the steering (top) and acceleration (bottom) trajectories for ADMM-NNMPC and NNMPC solutions in the two-lane (left) and three-lane (right) scenarios. Since ADMM-NNMPC solves for the optimal solution, it actively interacts with the other vehicles to open a gap for itself to merge into. Therefore, the steering trajectory in ADMM-NNMPC is more aggressive that the NNMPC. Lastly, the acceleration gradually changes in ADMM-NNMPC to reach the desired speed while minimizing jerk.  

\subsection{Limitations and Future Works}\label{Sec:Limitations}
% Our method is only guaranteed to be locally optimal due to highly non-convex nature of the neural networks. \bae{I doubt that it is a limitation of our proposal. It is a ``challenge'' associate to the nature of the problem. Although decomposed, the problems are still complex and their scalability could be the main limitations.}
Although we reduce the problem complexity by decomposing it into smaller sub-problems, these sub-problems are still complex which makes the approach non-scalable. 
Furthermore, due to the large neural network size and re-computation of gradients at each iteration, our current implementation runs slower than real-time. Nevertheless, having a slow offline optimization is useful, as it can serve as a benchmark when developing faster heuristic methods, ideally, we would like to increase the efficiency. Our approach can be made faster by training another neural network to estimate the original neural network's gradients and developing faster optimization libraries. Thus, future works include: (i) designing a smaller network trained with knowledge distillation~\cite{mirzadeh2020improved},
% \bae{cite papers on distillations},
or (ii) expediting neural network's gradient estimation using an offline-trained function approximator such as a neural network. 
% \bae{I am not familiar with papers without Conclusion section. Is it a common practice?}
% \di{Because of the highly non-convex nature of neural networks our solution is only guaranteed to be locally optimal.
% Because of XXX our current implementation runs slower than realtime at approximately XX Hz.
% }

% \di{remove or shorten conclusion for space}
% \begin{comment}
\section{Conclusions}\label{Sec:Conclusion}
With the importance of motion planning strategies being interaction-aware, e.g., lane changing in dense traffic for autonomous vehicles, this paper investigates mathematical solutions of a model predictive control with a neural network that estimates interactive behaviors. The problem is highly complex due to the non-convexity of the neural network, and we show that the problem can be effectively solved by decomposing it into sub-problems by leveraging the alternating direction method of multipliers (ADMM). This paper further examines the convergence of ADMM in presence of the neural network, which is one of the first attempts in the literature. The simple numerical study supports the provably optimal solutions being effective. The computational burden due to the complexity is still a limitation, and improving the computation efficiency remains for future work. That said, having a provably optimal solution is valuable as a benchmark when developing heuristic methods.

\begin{comment}

We propose an MPC-based motion planner that incorporates AV’s decision and surrounding vehicles’ interactive behaviors into safety constraints to perform complex maneuvers.
% We design an interaction-aware motion planner that interacts with
% other vehicles to perform complex maneuvers in a locally-optimal manner. 
Our planner utilizes a neural network-based interactive trajectory predictor and analytically integrates it with the MPC. In this paper, we solve the optimal control problem of MPC containing the non-convex neural network predictions using the ADMM algorithm. We also show the convergence of the ADMM algorithm in non-convex optimization and provide sufficient conditions on the neural network for convergence of the ADMM algorithm. We provide an empirical study by comparing our method with a baseline heuristic method.

There  are  several  possible  avenues  for  future  research. It is of interest to pre-compute the gradients of the neural network offline to make the optimizations faster. Designing a smaller neural network and using knowledge distillation to train it using the pre-trained neural network remains an important avenue for future work. Designing a smaller network with a small Lipschitz constant can improve the convergence rate of the algorithm and make the algorithm faster for real-time applications. 
\end{comment}

\appendix

\subsection{Proof of Lemma~\ref{lemma: feasibility}}\label{feasibility proof Appendix}
$C$ in~\eqref{eq:linear_equality_coeff} is a lower triangular matrix with diagonal entries as $-1$. Hence, $C$ is a full rank matrix of  rank $4T_p$, and $\text{Im}(C) =\mathbb{R}^{4T_p}$.  We have, $\text{Im}(Q) = \{ y \in \mathbb{R}^{4T_p}| \ y = Qx = [A, B]x \ \text{such that } x \in \mathbb{R}^{2T_p} \} \subseteq \mathbb{R}^{4T_p} = \text{Im}(C)$. $\hfill \blacksquare$

\subsection{Proof of Lemma~\ref{lemma:Lipschitz sub-minimization paths}}\label{Lipschitz sub_minimization proof Appendix}

$A$ and $B$ are full column rank matrices of column rank $T_p$. Furthermore, $C$ is a full rank matrix of rank $4T_p$. Therefore, their null spaces are trivial, and hence, $H_1, H_2, H_3$ reduces to linear operators and satisfies the Lemma.  $\hfill \blacksquare$

\subsection{Proof of Lemma~\ref{Lemma: Lipschitz smoothness}}\label{Lipschitz smoothness proof Appendix}

$\Phi_1(\boldsymbol{\Delta})$, $\Phi_2(\boldsymbol{\alpha})$, and $\Phi_3(\boldsymbol{Z})$ are $C^2$ functions, 
% twice-differentiable convex functions on the compact sets $\mathcal{D}$, $\mathcal{A}$, $\mathcal{Z}$, respectively, 
and hence, Lipschitz differentiable. Therefore, to show the Lipschitz differentiability of 
 $J$, it is sufficient to show that $b_i(\boldsymbol{Z})$, $i \in \mathcal{V}$, is Lipschitz differentiable for any $\tau \in \{1, \ldots, T_p\}$. For brevity of space, we define our notations in terms of $w \in \{x,y\}$ where $w$ can either be $x$ or $y$. 
 Let $q_w(\tau) :=2(w(\tau)-\phi_{i,w}(\tau-1))$.
% and $q_y(\tau) := 2(y(\tau)-\phi_{i,y}^{\boldsymbol{Z}}(\tau-1))$.
We have
\begin{align*}
        \frac{\partial b_i(\boldsymbol{Z})}{\partial x(k)} = 
        \begin{cases}
        -q_x(\tau)\frac{\partial \phi_{i,x}(\tau-1)}{\partial x(k)} - \\ \quad  q_y(\tau)
       \frac{\partial \phi_{i,y}(\tau-1)}{\partial x(k)}, \  \text{for} \ k \le \tau-1 \\
        q_x(\tau),  \ \ \ \ \text{for} \ k=\tau\\
        0, \  \ \ \  \text{for} \ k \in \{\tau+1, \ldots, T_p\}. 
       \end{cases}
\end{align*}
% \begin{align*}     
%       \frac{\partial b_i(\boldsymbol{Z})}{\partial y(k)} = 
%       \begin{cases}
%         -q_x(\tau)\frac{\partial \phi_{i,x}^{\boldsymbol{Z}}(\tau-1)}{\partial y(k)} - \\ \quad  q_y(\tau)
%       \frac{\partial \phi_{i,y}^{\boldsymbol{Z}}(\tau-1)}{\partial y(k)}, \  \text{for} \  k \le \tau-1 \\
%         q_y(\tau),  \ \ \ \ \text{for} \ k=\tau\\
%         0, \  \ \ \  \text{for} \ k \in \{\tau+1, \ldots, T_p\}. 
%       \end{cases}
% \end{align*}

Let $T^w_k:=\left|\frac{\partial b_i(\boldsymbol{Z}_1)}{\partial w(k)} - \frac{\partial b_i(\boldsymbol{Z}_2)}{\partial w(k)} \right|$ 
% and $T^y_k:=\left|\frac{\partial b_i(\boldsymbol{Z}_1)}{\partial y(k)} - \frac{\partial b_i(\boldsymbol{Z}_2)}{\partial y(k)} \right|$ 
for some $\boldsymbol{Z}_1, \boldsymbol{Z}_2 \in \mathcal{Z}$, and let $(x^m(\tau), y^m(\tau))$ 
% and  $(x^2(\tau), y^2(\tau))$ denote the $(x,y)$
denote the ego vehicle positions in $\boldsymbol{Z}_m$, where $m \in \{1,2\}$. Let $\phi_{i,w}^{\boldsymbol{Z}_m}$ denote $\phi_{i,w}$ corresponding to $\boldsymbol{Z}_m$. 
% Recall from (A2) that the gradient of the outputs of the neural network exists and is bounded. Therefore, 
Using assumption (A2) and mean-value theorem~\cite{rudin1976principles}, the neural network's outputs are Lipschitz continuous, i.e., $\| \phi_{i,w}^{\boldsymbol{Z}_1} - \phi_{i,w}^{\boldsymbol{Z}_2} \| \le \theta_w||\boldsymbol{Z}_1 - \boldsymbol{Z}_2||$.
% and $\| \phi_{i,y}^{\boldsymbol{Z}_1} - \phi_{i,y}^{\boldsymbol{Z}_2} \| \le \theta_y||\boldsymbol{Z}_1 - \boldsymbol{Z}_2||$. 
Let $\Delta w (\tau) = |w^1(\tau)-w^2(\tau)|$,  $\varphi_w(\tau-1) = | \phi_{i,w}^{\boldsymbol{Z}_2}(\tau-1) - \phi_{i,w}^{\boldsymbol{Z}_1}(\tau-1)  |$, and $\nu^w_x(\tau-1) = \Bigg|  \frac{\partial \phi_{i,w}^{\boldsymbol{Z}_1}(\tau-1)}{\partial x(k)}   - \frac{\partial \phi_{i,w}^{\boldsymbol{Z}_2}(\tau-1)}{\partial x(k)}  \Bigg|$.
% , and $\nu^y_x(\tau-1) = \Bigg|  \frac{\partial \phi_{i,y}^{\boldsymbol{Z}_1}(\tau-1)}{\partial x(k)}   - \frac{\partial \phi_{i,y}^{\boldsymbol{Z}_2}(\tau-1)}{\partial x(k)}  \Bigg|$. 
For any $k \in \{1, \ldots, \tau-1\}$:
\begin{align*}
   T_k^x
%   &= 2\Bigg| x^1(\tau)\frac{\partial \phi_{i,x}^{\boldsymbol{Z}_1}(\tau-1)}{\partial x(k)}   - x^2(\tau)\frac{\partial \phi_{i,x}^{\boldsymbol{Z}_2}(\tau-1)}{\partial x(k)} + \phi_{i,x}^{\boldsymbol{Z}_2}(\tau-1)\frac{\partial \phi_{i,x}^{\boldsymbol{Z}_2}(\tau-1)}{\partial x(k)}-\phi_{i,x}^{\boldsymbol{Z}_1}(\tau-1)\frac{\partial \phi_{i,x}^{\boldsymbol{Z}_1}(\tau-1)}{\partial x(k)} + \\
%   & \quad \quad \quad y^1(\tau)\frac{\partial \phi_{i,y}^{\boldsymbol{Z}_1}(\tau-1)}{\partial x(k)}   - y^2(\tau)\frac{\partial \phi_{i,y}^{\boldsymbol{Z}_2}(\tau-1)}{\partial x(k)} + \phi_{i,y}^{\boldsymbol{Z}_2}(\tau-1)\frac{\partial \phi_{i,y}^{\boldsymbol{Z}_2}(\tau-1)}{\partial x(k)}-\phi_{i,y}^{\boldsymbol{Z}_1}(\tau-1)\frac{\partial \phi_{i,y}^{\boldsymbol{Z}_1}(\tau-1)}{\partial x(k)} 
%   \Bigg| \\
   &\le 2 \Delta x (\tau)\Bigg|\frac{\partial \phi_{i,x}^{\boldsymbol{Z}_1}(\tau-1)}{\partial x(k)}   \Bigg| + 2 \Delta y (\tau)\Bigg|\frac{\partial \phi_{i,y}^{\boldsymbol{Z}_1}(\tau-1)}{\partial x(k)}   \Bigg| + \\
   & \quad  \quad  2|x^2(\tau)|\nu^x_x(\tau-1)   +   2\varphi_x(\tau-1)\Bigg|    \frac{\partial \phi_{i,x}^{\boldsymbol{Z}_2}(\tau-1)}{\partial x(k)}  \Bigg| + \\
  & \quad \quad 2|y^2(\tau)|\nu^y_x(\tau-1)  + 2 \varphi_y(\tau-1) \Bigg|    \frac{\partial \phi_{i,y}^{\boldsymbol{Z}_2}(\tau-1)}{\partial x(k)}  \Bigg| + \\
  & \quad \quad 2| \phi_{i,x}^{\boldsymbol{Z}_1}(\tau-1) |\nu^x_x(\tau-1) +  2| \phi_{i,y}^{\boldsymbol{Z}_1}(\tau-1) |\nu^y_x(\tau-1) \\
   &\le 2\theta_x\Delta x (\tau) + 2x_{max}  \nu^x_x(\tau-1)+ 2\theta_x\varphi_x(\tau-1) +  \\
   & \quad \quad 2s_x\nu^x_x(\tau-1)+ 2\theta_y\Delta y (\tau) + 2y_{max}\nu^y_x(\tau-1) + \\
   & \quad \quad 2\theta_y \varphi_y(\tau-1) + 2s_y\nu^y_x(\tau-1) \\
%   &\le 2(\theta_x(1+ \theta_x)+\theta_y(1+\theta_y) + \\
%   &  \quad \quad \quad  (x_{max} + y_{max} + s_x + s_y)L_{\nabla\phi} )\|\boldsymbol{Z}_1 - \boldsymbol{Z}_2\| \\
   &=L_1\|\boldsymbol{Z}_1 - \boldsymbol{Z}_2\|,
\end{align*}
where $L_1:=2(\theta_x(1+ \theta_x)+\theta_y(1+\theta_y) + (x_{max} + y_{max} + s_x + s_y)L_{\nabla\phi} )$, $x_{\max}$ and $y_{\max}$ are the bounds on the ego vehicle's $x$ and $y$ coordinates, respectively.

Similarly, for $k=\tau$, we have:
\begin{align*}
   T_k^x
%   &=2|x^2(\tau) - x^1(\tau) + \phi_{i,x}^{\boldsymbol{Z}_1}(\tau-1)- \phi_{i,x}^{\boldsymbol{Z}_2}(\tau-1)| \\
   &\le 2|x^2(\tau) - x^1(\tau)| + 2|\phi_{i,x}^{\boldsymbol{Z}_1}(\tau-1)- \phi_{i,x}^{\boldsymbol{Z}_2}(\tau-1)| \\
   & \le L_2||\boldsymbol{Z}_1 - \boldsymbol{Z}_2||,
\end{align*}
where $L_2 = 2(1+ \theta_x)$.

Similarly, $T^y_k \le L_1||\boldsymbol{Z}_1-\boldsymbol{Z}_2||$ for any $k \in \{0, \ldots, \tau-1\}$, and $T^y_k \le L_3||\boldsymbol{Z}_1-\boldsymbol{Z}_2||$, where $L_3= 2(1+\theta_y)$, for $k=\tau$.

Therefore, $ ||\nabla b_i(\boldsymbol{Z}_1)- \nabla b_i(\boldsymbol{Z}_2)|| \le L_g||\boldsymbol{Z}_1-\boldsymbol{Z}_2||$, where $L_g = T_p(\max\{L_1, L_2\} + \max \{L_1, L_3 \})$. Hence, $J(\boldsymbol{\Delta}, \boldsymbol{\alpha}, \boldsymbol{Z})$ in~\eqref{eq:compact_obj_final} is Lipschitz differentiable.
$\hfill \blacksquare$

\subsection{Proof of Theorem~\ref{thm1}}\label{convergence theorem proof appendix}
Since $C$ is a full rank matrix, $Im(C) = \mathbb{R}^{4T_p}$, and hence, $D \in Im(C)$. Recall that the feasible sets for $\boldsymbol{\Delta}$, $\boldsymbol{\alpha}$, and $\boldsymbol{Z}$ are bounded, i.e.,  $\boldsymbol{\Delta} \in\mathcal{D}, \boldsymbol{\alpha} \in \mathcal{A}$, and $\boldsymbol{Z}\in\mathcal{Z}$.
Using these results and Lemmas~\ref{lemma: feasibility}-\ref{Lemma: Lipschitz smoothness}, the optimization problem satisfies all the assumptions required for convergence of ADMM in non-convex and non-smooth optimization~\cite{wang2019global}. Utilizing~\cite[Theorem 2]{wang2019global} proves the convergence of Algorithm~\ref{alg:mpc_admm} for any sufficiently large $\rho  > \max \{1, (1 + 2\sigma_{\min}(C))L_{J}M\}$. $\hfill \blacksquare$

\bibliographystyle{IEEEtran}
\bibliography{ref}
\end{document}